\pdfoutput=1
\documentclass[11pt]{article}

\usepackage[final]{acl}
\newcommand{\STAB}[1]{\begin{tabular}{@{}c@{}}#1\end{tabular}}
\usepackage{times}
\usepackage{latexsym}
\usepackage{amsmath}
\usepackage{xcolor}
\usepackage{anyfontsize}
\usepackage{multirow}
\usepackage[T1]{fontenc}
\usepackage{enumitem}
\usepackage{booktabs}
\usepackage[utf8]{inputenc}
\usepackage{hyperref}
\usepackage{microtype}

\usepackage{inconsolata}
\usepackage{graphicx}

\usepackage{tikz}
\usepackage{tikz-dependency}

\title{Evaluation of Multilingual Image Captioning: \\How far can we get with CLIP models?}

\author{
  \textbf{Gonçalo Gomes\textsuperscript{1,2}},
  \textbf{Chrysoula Zerva\textsuperscript{1,3}},
  \textbf{Bruno Martins\textsuperscript{1,2}}
\\
  \textsuperscript{1}Instituto Superior Técnico, University of Lisbon
\\
  \textsuperscript{2}INESC-ID
\\
  \textsuperscript{3}Instituto de Telecomunicações
\\
  \small{
    {\{goncaloecgomes, chrysoula.zerva, bruno.g.martins\}@tecnico.ulisboa.pt}
  }
}

\begin{document}
\maketitle
\begin{abstract}
The evaluation of image captions, looking at both linguistic fluency and semantic correspondence to visual contents, has witnessed a significant effort. Still, despite advancements such as the CLIPScore metric, multilingual captioning evaluation has remained relatively unexplored. This work presents several strategies, and extensive experiments, related to evaluating CLIPScore variants in multilingual settings. To address the lack of multilingual test data, we consider two different strategies: (1) using quality aware machine-translated datasets with human judgements, and (2) re-purposing multilingual datasets that target semantic inference and reasoning. Our results highlight the potential of finetuned multilingual models to generalize across languages and to handle complex linguistic challenges. Tests with machine-translated data show that multilingual CLIPScore models can maintain a high correlation with human judgements across different languages, and additional tests with natively multilingual and multicultural data further attest to the high-quality assessments. 
\end{abstract}

\section{Introduction}

Computer-generated image captions are nowadays commonly used as descriptive annotations. The Image Captioning (IC) task has been extensively studied, including in multilingual settings, with many recent approaches combining established vision encoders with large language model decoders~\cite{ramos2023smallcap, ramos2023retrieval, ramos2023lmcap,yang2023multicapclip,geigle2023mblip,ramos2024mpaella}. The automatic evaluation of captions, accounting for linguistic fluency and alignment with visual contents, has also witnessed a significant effort. Approaches like CLIPScore~\cite{hessel2021clipscore} have been proposed to evaluate captions through cosine similarity between image and text embeddings, leveraging large-scale pre-trained vision-and-language models and achieving high correlations with human judgments. Still, despite the many recent advancements, most approaches are English-centric, while multilingual image captioning evaluation has remained relatively unexplored, and lacking in resources. 

This work explores the use of CLIPScore in multilingual captioning evaluation. Given the lack of available benchmarks for the evaluation of multilingual captioning metrics, we propose two different evaluation strategies: (1) using a combination of Machine Translation (MT) and Quality Estimation (QE) models to obtain high-quality multilingual data from English-centric benchmarks, and (2) re-purposing multilingual benchmarks originally designed for evaluation of the semantic inference and reasoning capabilities of vision-language models. 

Through extensive experiments, we show that multilingual CLIP models achieve comparable or even better performance on English benchmarks, while allowing for multilingual assessments. We also propose a multilingual finetuning strategy for CLIPScore, that allows to account for linguistic and cultural diversity while learning from human judgements, resulting in further performance improvements.  Performance generally increases according to model size, and larger models, trained on more data, attained similar or even better performance to methods that extended the original CLIPScore~\cite{sarto2023positive,kim2022mutual,hu2023infometic,kim2023pr,narins2024validated,wada2024polos}. Tests with machine-translated data show that multilingual CLIPScore can also maintain a high correlation with human judgements across different languages, and additional tests with natively multilingual and multicultural data further attest to high-quality assessments across languages.

Our study highlights the importance of multilingual and multicultural research in the evaluation of image captions, aiming to inspire the development of multilingual frameworks in this domain. We show that multilingual models, trained on equally sized datasets, perform just as well as English-only models on English assessments, while also excelling across various evaluation tasks that involve multilingual and multicultural data, making them far more versatile and valuable assets than the corresponding English-centric models.

In sum, our main contributions include (i) a finetuning strategy that accounts for linguistic and cultural diversity, as well as alignment with human judgements (Section \ref{sec: multilingual_clipscore}); (ii) an MT-based extension of English-centric benchmarks to multiple languages, incorporating human evaluations and diverse linguistic phenomena, while preserving the original benchmarks' quality (Sections \ref{sec: multilingual_correlation} and \ref{sec: multilingual_classification}) ; and (iii) an adaptation of existing multilingual and multicultural datasets for captioning metric evaluation (Section \ref{sec: multilingual_classification}). The code and adapted datasets supporting our evaluation experiments are available in a public GitHub repository\footnote{\url{https://github.com/gecgomes/Multilingual_IC_Eval}}.

\section{Related Work}

Conventional image captioning evaluation has relied on reference-based assessments, where machine-generated captions are compared against human-generated ones (i.e., the references). 
Frequently used metrics such as BLEU or CIDEr~\cite{vedantam2015cider}  rely on lexical matches, and hence may fail to capture fine nuances and semantic overlaps in rich captions. A recent shift in the evaluation paradigm involves the use of learned vision-and-language models to enable evaluation through reference-free metrics.

The CLIPScore metric~\cite{hessel2021clipscore} was one of the first proposals for evaluating image captions that departed from traditional metrics. Grounded in a vision-and-language encoder, specifically the original Contrastive Language-Image Pretraining (CLIP) model~\cite{radford2021learning}, this strategy employs a modified cosine similarity between representations for the input image and the caption under evaluation. CLIPScore exhibits a high correlation with human judgments across various datasets, and despite being a reference-free metric, it even surpasses established reference-based metrics like BLEU and CIDEr. Today, CLIPScore-based metrics are widely used for evaluating image captions, and have inspired the development of numerous new evaluation metrics that build on CLIP~\cite{sarto2023positive,hu2023infometic,kim2022mutual}, including reference-based variants.

Despite the advancements, recent studies have noted that CLIPScore can lack rating granularity, emphasising the need for better benchmark datasets to evaluate image captioning metrics~\cite{ahmadi2024examination}. Datasets like VICR~\cite{narins2024validated} or Polaris~\cite{wada2024polos} have employed more rigorous methods for collecting human ratings, but remain limited to English. Alongside these datasets, researchers have shown that models specifically trained for image captioning evaluation can slightly outperform CLIPScore. We corroborate these findings by proposing a finetuned version of CLIPScore that surpasses previous variants.

Recent studies emphasize the importance of evaluating visio-linguistic grounding capabilities with more complex linguistic constructs \cite{parcalabescu2022valse}. Additionally, the absence of multilingual and multicultural benchmarks has been a key topic of discussion. For instance, \citet{kim2023pr} proposed a multilingual image captioning evaluation method. They finetuned the CLIP text encoder using a language-agnostic approach to distinguish between original and perturbed text. Moreover, the authors introduced a new dataset aimed at evaluating multilingual captioning metrics, although it has not yet been made publicly available.

The lack of linguistic and cultural diversity in vision-language datasets has been further criticized by~\citet{liu2021visually}, who argued that relying on English-based lexical databases and image queries introduces a North American and Western European bias, since existing vision-linguistic datasets often fail to capture concepts relevant to non-English languages and cultures outside of Europe and North America. To mitigate this limitation, the authors built MaRVL, which proposes a new protocol for constructing an ImageNet-style class hierarchy that better reflects diverse languages and cultures, together with textual descriptions reflecting these classes. They benchmarked several multilingual multimodal models, but these often performed just above chance, highlighting the challenges of handling out-of-distribution concepts and images. The results call for a reassessment of model robustness and accuracy, and for the development of more inclusive systems.

Similarly, \citet{bugliarello2022iglue} highlighted the importance of multilingual benchmarks, noting that vision-and-language research has primarily focused on English tasks. To address this, the authors proposed a multilingual dataset for evaluating vision-language inference, where models predict entailment relationships between a textual hypothesis and an image premise, expanding the scope of evaluation to include diverse languages.

\section{Multilingual CLIPScore} \label{sec: multilingual_clipscore}
The CLIPScore metric uses an adjusted cosine similarity to compare representations for the input image and the caption being assessed, as originally described by \citet{hessel2021clipscore}. In our work, we adopted the original formulation. A more detailed explanation of the CLIPScore and RefCLIPScore metrics can be found in Appendix~\ref{app:clipscore}.

To boost performance across languages, we propose a strategy to finetune multilingual CLIP models in a setting that considers both linguistic and cultural diversity, while accounting for human preference alignment. Two distinct datasets were used for finetuning. The first, CrossModal-3600, focuses on multilingual captions and multicultural imagery \cite{thapliyal2022crossmodal}, whereas the second, VICR, comprises English image-caption pairs that are evaluated by humans \cite{narins2024validated}, and which we machine translated to different languages following a strict quality-aware translation scheme to help maintain high quality. We also propose the combination of different training objectives tailored to the specific characteristics of each of the two datasets.

In more detail, to enhance the model's ability to process multilingual and multicultural instances, we finetuned it on both datasets using the original CLIP contrastive loss, which can be formally described as follows:
\begin{equation}
\resizebox{0.95\columnwidth}{!}{%
    $\mathcal{L}_C = -\frac{1}{2N}\sum_{i=1}^N\left[\text{log}\frac{e^{s_{i,i}/\tau}}{\sum^N_{j=1} e^{s_{i,j}/\tau}} + \text{log}\frac{e^{s_{i,i}/\tau}}{\sum^N_{j=1} e^{s_{j,i}/\tau}}\right].$
}
\end{equation}
In the equation, $N$ is the number of image-text pairs in a batch, $s_{i,j}$ is the similarity score between the \textit{i}-th image and the \textit{j}-th text description, and $\tau$ is a temperature parameter that scales the similarity scores and helps in controlling the concentration level for the distribution of scores.

For the second dataset, to improve the alignment of CLIPScore with human ratings, we also considered a Pearson correlation loss:
\begin{equation}
\resizebox{0.65\columnwidth}{!}{%
    $\mathcal{L}_P = 1 - \frac{(x-\overline{x})^T(y-\overline{y})}{||(x-\overline{x})||\cdot ||(y-\overline{y})||},$
    }
\end{equation}
where $x$ is the vector of CLIPScore values, $y$ is a vector with the human rating scores, and $\overline{x}$ and $\overline{y}$ are the respective average values. 

Considering that both loss functions can benefit from larger batch sizes, we sample instances for training by alternating between each task, without mixing instances with respect to different losses in the same batch. We accumulate gradients for two steps before updating the network, effectively combining both loss effects while leveraging the benefits of larger batches, i.e., $\mathcal{L} \sim \mathcal{L}_C + \mathcal{L}_P$.

\section{Experimental Evaluation}

This section presents the datasets, the experimental setup, and the results for different CLIP models, considering English, multilingual, and multicultural scenarios for image captioning evaluation.

\subsection{Datasets}

To ensure a fair and comprehensive evaluation of our multilingual models, we extended several well-established English-centric datasets to a multilingual scenario. These datasets, originally developed for assessing human judgments in image-captioning tasks, contain one or more human quality assessments for each image-caption pair.

\begin{itemize}\setlength\itemsep{-0.3em} 
    \item \textbf{Expert} (Flickr8K-Expert), consisting of $5,664$ pairs~\cite{hodosh2013framing}.
    \item \textbf{Crowdflower} (Flickr8K-CF), with a total of $47,830$ pairs~\cite{hodosh2013framing}. 
    \item \textbf{Composite}, containing with a total of $13,146$ pairs~\cite{aditya2015images}.
    \item \textbf{VICR}, with $10,175$ training, $2,310$ validation, and $3,161$ test pairs~\cite{narins2024validated}. 
\end{itemize}

In addition to using the aforementioned datasets featuring human ratings, we also evaluated the robustness of our models to various linguistic phenomena using the \textbf{VALSE} dataset~\cite{parcalabescu2022valse}, which is used to perform evaluation through a binary classification task, and which comprises $6,704$ correct image-caption pairs alongside their foil caption versions.

While the comparison between multilingual and monolingual models on English data offers some insights, it provides a limited perspective on multilingual performance. Unfortunately, high-quality multilingual resources featuring curated human assessments of caption quality are scarce or non-existent, restricting the scope of multilingual evaluation. To overcome this, we developed a translation scheme that leverages large machine translation models~\cite{m2m100,alves2024tower,liu2020multilingual}, paired with language and translation quality estimation models~\cite{rei2022cometkiwi,rei2023scaling}, to translate English captions with pre-existing human assessments into multiple languages. This approach targets high translation quality, {filtering out low-quality translations and thus ensuring the validity of human judgments across target languages}. Further details about our translation scheme can be found in Appendix \ref{app:translation}.

Our language selection is in line with recent machine translation studies~\cite{alves2024tower}, covering high-resource languages (i.e., English, French, German, Spanish, and Chinese) and also mid- (i.e., Portuguese, Italian, and Russian) to low-resource languages (i.e., Dutch and Korean). We translated both the \textbf{VICR} and \textbf{VALSE} datasets into the nine aforementioned languages using our MT scheme. 

In addition, to further expand our multilingual evaluation, we used {natively multilingual and multicultural datasets}, i.e., \textbf{XVNLI}~\cite{bugliarello2022iglue} and \textbf{MaRVL}~\cite{liu2021visually}, {re-purposing them into classification tasks for the evaluation of image captioning metrics}. 
We also expanded both native multilingual datasets by translating them into English, allowing us to compare a multilingual model with its English-only counterpart.


\subsection{Evaluation Metrics}

We evaluate the different models using correlation with human judgements, and also through classification tasks. Regarding the correlation experiments, we measure performance using three different correlation coefficients, namely Spearman $\rho$ and Kendall $\tau$ with variations $b$ and $c$. The correlation metrics are formally defined in Appendix~\ref{app:metrics}. 

For the classification experiments, we measure accuracy under the assumption that a caption entailed by an image should reflect a higher CLIPScore than a contradiction/foil caption. 

\subsection{Experiments and Results}

This section presents experimental results for the different models and evaluation datasets, establishing a comparison with previously reported results and contributing to the multi-linguistic exploration of existing models and datasets. We also performed a qualitative study focusing on image-caption pairs that feature concepts that could be associated with cultural bias, which is reported in Appendix~\ref{app:qualitative}. 

\subsubsection{Model Selection} \label{sec:model_selection}

\begin{table*}[t!]
\renewcommand{\arraystretch}{1.4}
\centering
\fontsize{20pt}{20pt}\selectfont
\resizebox{2\columnwidth}{!}{%
\begin{tabular}{c l c c c c c c c c c c c c c c c c c c}
&  & \multicolumn{2}{c}{Size (B)} & ~ & \multicolumn{3}{c}{VICR} & ~ & \multicolumn{3}{c}{Expert} & ~ & \multicolumn{3}{c}{CrowdFlower} & ~ & \multicolumn{3}{c}{Composite} \\  \cline{3-4} \cline{6-8} \cline{10-12} \cline{14-16} \cline{18-20} 

 & & Model & Data & & $\tau_b$ & $\tau_c$ & $\rho$ & & $\tau_b$ & $\tau_c$ & $\rho$ & & $\tau_b$ & $\tau_c$ & $\rho$ & & $\tau_b$ & $\tau_c$ & $\rho$ \\ \hline 
\multicolumn{1}{l}{\STAB{\multirow{10}{*}{\rotatebox[origin=c]{90}{English}}}} & \href{https://huggingface.co/openai/clip-vit-base-patch32}{openai/clip-vit-base-patch32} & 0.1 & 0.4  & & 60.7 & 67.1 & 76.9 & & 51.1 & 51.5 & 63.1 & & 34.4 & 17.8 & 42.4 & & 50.6 & 51.4 & 67.9  \\ \cline{2-20}
\multicolumn{1}{l}{} & \href{https://huggingface.co/apple/DFN5B-CLIP-ViT-H-14-378/tree/main}{apple/DFN5B-CLIP-ViT-H-14-378}~~~ & 0.9 & 5.0 & & 67.4 & \textbf{74.4} & \textbf{83.1} & & 56.3 & \textbf{56.6} & \textbf{68.4} & & \textbf{38.5} & \textbf{19.9} & \textbf{47.1} & & 55.0 & 55.9 & 72.8  \\
\multicolumn{1}{l}{}& \href{https://huggingface.co/apple/DFN5B-CLIP-ViT-H-14}{apple/DFN5B-CLIP-ViT-H-14}      & 0.9 & 5.0 & &  66.5 & 73.5 & 82.4 & & 55.6 & 56.0 & 67.7 & & 38.2 & 19.7 & 46.8 & & 54.5 & 55.3 & 72.3  \\
\multicolumn{1}{l}{}&\href{https://huggingface.co/apple/DFN2B-CLIP-ViT-L-14}{apple/DFN2B-CLIP-ViT-L-14}     & 0.4 & 2.0 & &  65.8 & 72.6 & 81.6 & & 55.5 & 55.6 & 67.5 & & 37.1 & 19.2 & 45.6 & & 53.6 & 54.3 & 71.3  \\
\multicolumn{1}{l}{}&\href{https://huggingface.co/laion/CLIP-ViT-g-14-laion2B-s12B-b42K}{laion/CLIP-ViT-g-14-laion2B-s12B-b42K}     & 1.3 & 2.0  & & 66.4 & 73.0 & 82.0 & & 55.3 & 55.1 & 66.9 & & 37.1 & 19.1 & 45.5 & & 54.9 & 55.7 & 72.7 \\
\multicolumn{1}{l}{}&\href{https://huggingface.co/laion/CLIP-ViT-H-14-laion2B-s32B-b79K}{laion/CLIP-ViT-H-14-laion2B-s32B-b79K}   & 0.9 & 2.0    & & 66.4 & 73.1 & 82.1 & & 54.9 & 54.9 & 66.7 & & 37.2 & 19.2 & 45.6 & & 53.6 & 54.4 & 71.2  \\
\multicolumn{1}{l}{}&\href{https://huggingface.co/laion/CLIP-ViT-L-14-laion2B-s32B-b82K}{laion/CLIP-ViT-L-14-laion2B-s32B-b82K}  & 0.4 & 2.0     & & 65.9 & 72.6 & 81.7 & & 54.4 & 54.5 & 66.3 & & 36.7 & 18.9 & 45.1 & & 53.6 & 54.4 & 71.3  \\
\multicolumn{1}{l}{}&\href{https://huggingface.co/BAAI/AltCLIP}{BAAI/AltCLIP}        & 0.8 & - &  & 65.1 & 71.9 & 81.1 & & 54.1 & 54.4 & 66.2 & & 36.2 & 18.7 & 44.6 & & 53.8 & 54.6 & 71.6  \\
\multicolumn{1}{l}{}&\href{https://huggingface.co/openai/clip-vit-large-patch14-336}{openai/clip-vit-large-patch14-336}  & 0.4 & 0.4 &   & 62.0 & 68.5 & 78.1 & & 52.6 & 53.0 & 64.6 & & 35.4 & 18.3 & 43.7 & & 52.8 & 53.6 & 70.3  \\
\multicolumn{1}{l}{}&\href{https://huggingface.co/openai/clip-vit-large-patch14}{openai/clip-vit-large-patch14}   & 0.4 & 0.4   &     & 62.3 & 68.9 & 78.5 & & 52.6 & 53.0 & 64.6 & & 35.2 & 18.2 & 43.4 & & 52.5 & 53.3 & 70.0  \\ \hline
\multicolumn{1}{l}{\STAB{\multirow{10}{*}{\rotatebox[origin=c]{90}{Multilingual}}}}&\href{https://huggingface.co/laion/CLIP-ViT-H-14-frozen-xlm-roberta-large-laion5B-s13B-b90k}{laion/CLIP-ViT-H-14-frozen-xlm-roberta-large-laion5B-s13B-b90k}~~~~~~ & 1.1 & 5.0 && \textbf{67.6} & 73.0 & 82.4 & & \textbf{57.4} & 54.3 & 67.3 & & 38.2 & 19.4 & 46.2 & & \textbf{55.5} & \textbf{56.2} & \textbf{73.2}  \\
\multicolumn{1}{l}{}&\href{https://huggingface.co/BAAI/AltCLIP-m18}{BAAI/AltCLIP-m18}     & 1.1 & -   &  & 66.6 & 73.4 & 82.3 & & 54.8 & 55.0 & 66.8 & & 37.1 & 19.2 & 45.6 & & 55.2 & 56.0 & 73.0  \\
\multicolumn{1}{l}{}&\href{https://huggingface.co/M-CLIP/XLM-Roberta-Large-Vit-B-16Plus}{M-CLIP/XLM-Roberta-Large-Vit-B-16Plus}  & 0.7 & 0.4   &    & 64.6 & 71.3 & 80.6 & & 54.6 & 54.8 & 66.7 & & 36.0 & 18.6 & 44.3 & & 52.1 & 52.9 & 69.6 \\
\multicolumn{1}{l}{}&\href{https://huggingface.co/BAAI/AltCLIP-m9}{BAAI/AltCLIP-m9}                                   & 0.8 & -            &         & 64.2 & 70.9 & 80.3 & & 54.1 & 54.4 & 66.2 & & 36.4 & 18.8 & 44.8 & & 53.9 & 54.6 & 71.6 \\
\multicolumn{1}{l}{}&\href{https://huggingface.co/jinaai/jina-clip-v1}{jinaai/jina-clip-v1}~~~~~~ & 0.3 & - & & 63.9 & 70.2 & 79.7 & & 54.2 & 53.6 & 65.6 & & 35.0 & 18.0 & 43.0 & & 53.2 & 54.1 & 70.7 \\
\multicolumn{1}{l}{}&\href{https://huggingface.co/laion/CLIP-ViT-B-32-xlm-roberta-base-laion5B-s13B-b90k}{laion/CLIP-ViT-B-32-xlm-roberta-base-laion5B-s13B-b90k}~~~~~~ & 0.3 & 5.0 & & 63.3 & 69.8 & 79.3 & & 52.7 & 52.8 & 64.5 & & 35.2 & 18.2 & 43.3 & & 49.9 & 50.6 & 67.1 \\
\multicolumn{1}{l}{}&\href{https://huggingface.co/M-CLIP/XLM-Roberta-Large-Vit-L-14}{M-CLIP/XLM-Roberta-Large-Vit-L-14}          & 0.9 & 0.4        & & 62.2 & 68.7 & 78.4 & & 53.0 & 53.4 & 65.0 & & 35.4 & 18.3 & 43.7 & & 52.9 & 53.6 & 70.4 \\
\multicolumn{1}{l}{}&\href{https://huggingface.co/M-CLIP/XLM-Roberta-Large-Vit-B-32}{M-CLIP/XLM-Roberta-Large-Vit-B-32}           & 0.7 & 0.4      &  & 60.5 & 66.9 & 76.7 & & 51.8 & 52.2 & 63.9 & & 34.4 & 17.8 & 42.4 & & 50.7 & 51.4 & 67.9 \\
\multicolumn{1}{l}{}&\href{https://huggingface.co/sentence-transformers/clip-ViT-B-32-multilingual-v1}{sentence-transformers/clip-ViT-B-32-multilingual-v1} & 0.3 & 1.1 && 60.3 & 66.7 & 76.6 & & 51.5 & 51.8 & 63.6 & & 33.3 & 17.2 & 41.1 & & 48.9 & 49.6 & 65.9 \\ \hline
\end{tabular}
}
\vspace{-0.25cm}
\caption{Correlation between CLIPScore values and human rankings, considering a set of different English (top rows) and multilingual (bottom rows) CLIP models. The columns named "Size (B)" describe the size of the corresponding model in billions of parameters, and the amount of training data in billions of image-caption instances.}
\label{tab:english-only-corr}
\vspace{-0.25cm}
\end{table*}

To select the specific CLIP model to be used in our multilingual experiments, following the CLIPScore methodology described by \citet{hessel2021clipscore}, we first analysed correlation results between CLIPScore values and human ratings across four English datasets and considering publicly available CLIP models. Results are reported in Table \ref{tab:english-only-corr}, indicating that CLIPScore estimates improve significantly with larger CLIP models trained on more data, even if the training data is multilingual and the model is tested on English-centric data.

Apple's public ViT-H/14 model, with a $378^2$ pixel resolution and trained exclusively on English data, achieves the highest correlation results among the different English models. However, a multilingual model of the same size, trained on a mixture of English and multilingual data from LAION and with significantly less English-specific focus (i.e., \href{https://huggingface.co/laion/CLIP-ViT-H-14-frozen-xlm-roberta-large-laion5B-s13B-b90k}{laion/CLIP-ViT-H-14-frozen-xlm-roberta-large-laion5B-s13B-b90k}), emerges as the second-best alternative. 

Both models are comparable in size and were trained on 5 billion instances. While Apple's model relied solely on English data, the multilingual LAION model incorporated a diverse mixture of languages. Despite its reduced exposure to English-specific data, the multilingual model delivers competitive performance on English benchmarks, even surpassing Apple's model in several correlation metrics. This indicates that the multilingual training data enabled effective generalization, enhancing the model's capabilities even for tasks involving only English data.

We selected the \href{https://huggingface.co/laion/CLIP-ViT-H-14-frozen-xlm-roberta-large-laion5B-s13B-b90k}{laion/CLIP-ViT-H-14-frozen-xlm-roberta-large-laion5B-s13B-b90k} variant for further experiments, as it demonstrated the best performance. From this point forward, we will refer to this multilingual model as MLAION (ML), and to our finetuned version as MLAION-F (MLF).

Additionally, experiments detailed in Appendix~\ref{app:english_experiments} demonstrate that simply scaling up model size and training with more diverse data can yield performance on par with more complex captioning evaluation methods. Our study shows that using CLIPScore with better CLIP models can compete and even outperform specialized IC evaluation approaches. Our finetuned multilingual CLIP model outperformed the original multilingual LAION ViT-H and Apple's best model across all CLIPScore variants on well-established human judgment datasets. With references, it managed to surpass specialized architectures like VICR~\cite{narins2024validated}, Polos~\cite{wada2024polos}, InfoMetIC~\cite{hu2023infometic}, or RefPACScore~\cite{sarto2023positive}, achieving state-of-the-art performance comparable to CAMScore~\cite{cui2025evaluating}, G-VEval~\cite{tong2024g}, and CLAIR~\cite{chan2023clair}.

\subsubsection{Correlation on Multilingual Data} \label{sec: multilingual_correlation}

\begin{table*}[t!]
\centering
\renewcommand{\arraystretch}{1.15}
\resizebox{2\columnwidth}{!}{%
\begin{tabular}{l c c c   c   c c c   c   c c c   c   c c c }
 & \multicolumn{7}{c}{ Without Finetuning}  & ~ & \multicolumn{7}{c}{ With Finetuning}  \\ \cline{2-8}  \cline{10-16}
 & \multicolumn{3}{c}{ ML CLIPScore}       & ~ & \multicolumn{3}{c}{ ML RefCLIPScore} & ~ & \multicolumn{3}{c}{ MLF CLIPScore} & ~ & \multicolumn{3}{c}{ MLF RefCLIPScore} \\ 
 \cline{2-4} \cline{6-8} \cline{10-12} \cline{14-16}  
 & ~~$\tau_b / std$~~ & ~~$\tau_c / std$~~ & ~~$\rho / std$~~ &    & ~~$\tau_b / std$~~ & ~~$\tau_c / std$~~ & ~~$\rho / std$~~ &    & ~~$\tau_b / std$~~ & ~~$\tau_c / std$~~ & ~~$\rho / std$~~ &   & ~~$\tau_b / std$~~ & ~~$\tau_c / std$~~ & ~~$\rho / std$~~ \\   \hline 
English     &67.6 / 0.15 & 73.0 / 0.15 & 82.4 / 0.13&  &69.1 / 0.15 & 74.6 / 0.15 & 83.6 / 0.13&   &68.7 / 0.14 & 75.4 / 0.15 & 84.1 / 0.12&  &\textbf{69.7} / 0.14 & \textbf{76.5} / 0.15 & \textbf{84.8} / 0.12\\ \hline
German      &66.1 / 0.15 & 72.3 / 0.15 & 81.5 / 0.13&  &68.0 / 0.14 & 74.4 / 0.15 & 83.2 / 0.12&   &68.0 / 0.14 & 74.8 / 0.15 & 83.6 / 0.12&  &\textbf{69.1} / 0.14 & \textbf{75.9} / 0.14 & \textbf{84.4} / 0.12\\
French      &65.9 / 0.16 & 71.8 / 0.17 & 81.3 / 0.15&  &67.6 / 0.15 & 73.6 / 0.16 & 82.7 / 0.14&   &68.0 / 0.14 & 74.7 / 0.15 & 83.6 / 0.12&  &\textbf{69.1} / 0.14 & \textbf{75.9} / 0.15 & \textbf{84.4} / 0.12\\
Spanish     &66.3 / 0.15 & 72.7 / 0.16 & 81.8 / 0.14&  &67.9 / 0.14 & 74.4 / 0.15 & 83.2 / 0.12&   &67.8 / 0.14 & 74.5 / 0.15 & 83.5 / 0.12&  &\textbf{68.9} / 0.14 & \textbf{75.7} / 0.15 & \textbf{84.3} / 0.12\\
Chinese     &65.0 / 0.15 & 71.4 / 0.16 & 80.8 / 0.14&  &66.7 / 0.14 & 73.2 / 0.15 & 82.3 / 0.13&   &67.4 / 0.15 & 74.0 / 0.15 & 83.1 / 0.13&  &\textbf{68.5} / 0.14 & \textbf{75.2} / 0.15 & \textbf{83.9} / 0.12\\ 
Portuguese  &66.1 / 0.15 & 72.2 / 0.16 & 81.6 / 0.14&  &67.8 / 0.14 & 74.1 / 0.15 & 83.0 / 0.12&   &67.9 / 0.14 & 74.6 / 0.15 & 83.5 / 0.12&  &\textbf{69.0} / 0.14 & \textbf{75.8} / 0.14 & \textbf{84.3} / 0.12\\
Italian     &65.8 / 0.15 & 72.1 / 0.16 & 81.4 / 0.14&  &67.5 / 0.14 & 74.0 / 0.15 & 82.9 / 0.13&   &67.9 / 0.14 & 74.6 / 0.15 & 83.5 / 0.13&  &\textbf{68.9} / 0.14 & \textbf{75.7} / 0.15 & \textbf{84.3} / 0.12\\
Russian     &65.0 / 0.15 & 71.4 / 0.16 & 80.7 / 0.14&  &66.7 / 0.15 & 73.3 / 0.16 & 82.2 / 0.13&   &67.6 / 0.14 & 74.2 / 0.15 & 83.2 / 0.12&  &\textbf{68.8} / 0.14 & \textbf{75.6} / 0.14 & \textbf{84.2} / 0.12\\
Korean      &64.3 / 0.16 & 70.5 / 0.17 & 80.0 / 0.15&  &66.3 / 0.15 & 72.7 / 0.16 & 81.8 / 0.14&   &67.4 / 0.14 & 74.1 / 0.15 & 83.1 / 0.13&  &\textbf{68.3} / 0.14 & \textbf{75.0} / 0.15 & \textbf{83.8} / 0.12\\
Dutch       &65.8 / 0.14 & 72.0 / 0.15 & 81.4 / 0.13&  &67.7 / 0.14 & 74.1 / 0.15 & 83.0 / 0.12&   &68.2 / 0.14 & 74.9 / 0.14 & 83.8 / 0.12&  &\textbf{69.0} / 0.13 & \textbf{75.8} / 0.14 & \textbf{84.3} / 0.11\\ \hline 
Average~~   &65.8 / 0.15 & 71.9 / 0.16 & 81.3 / 0.14&  &67.4 / 0.14 & 73.7 / 0.15 & 82.7 / 0.13&   &67.9 / 0.14 & 74.6 / 0.15 & 83.5 / 0.12&  &\textbf{68.9} / 0.14 & \textbf{75.7} / 0.15 & \textbf{84.3} / 0.12\\
\hline 
\end{tabular}
}
\vspace{-0.25cm}
\caption{Correlation between multilingual CLIPScore values and human rankings, considering machine-translated versions of the VICR dataset into 9 different languages besides the original English. The last row presents macro-averaged correlation results across all the languages (including English). \textbf{Bold} values signify the best score per language. All observed score differences are statistically significant using the Williams test ($p < 0.01$). The std
values in each column represent the standard deviation in the test set for each metric across different models.  We used a stratified random sampling approach, performing 1,000 iterations to generate subsets containing 80\% of the original data, as we do not finetune all models.}
\label{tab:multilingual-results}
\vspace{-0.25cm}
\end{table*}

Table \ref{tab:multilingual-results} displays the correlation between multilingual CLIPScore values and human ratings across the different languages. Our finetuned version achieves significantly better correlations with human judgments, both in reference-based and reference-free settings, across all evaluated languages and correlation metrics. 
The finetuned CLIPScore model strongly correlates with human preferences in high-resource languages (i.e., English, French, German, Spanish, and Chinese), and it also exhibits equally strong performance in medium- and low-resource languages. Importantly, our finetuned CLIPScore model outperformed the pre-finetuned version even when the latter was using references, achieving a higher average correlation across all metrics without using any references. This observation is particularly useful as it encourages the application of the model to new instances, without requiring additional human input.

Appendix \ref{app:results} provides additional results regarding different model sizes and loss variants for model finetuning. The findings in the appendix support the idea that smaller CLIP models can obtain higher gains in correlation with human judgements when using our finetuning strategy, compared to the original models.

Delving deeper into the impact of MT quality, we note that in an ideal scenario, i.e. assuming perfect machine translation results and balanced CLIP performance across languages, the correlations between CLIPScore values across the different languages would equal one, signifying a perfect alignment. To explore deviations from this ideal behaviour, we use heatmaps to visually represent the interrelationships between CLIPScore values across the different languages.

\begin{figure*}[!t]
  \centering
  \includegraphics[width=2\columnwidth]{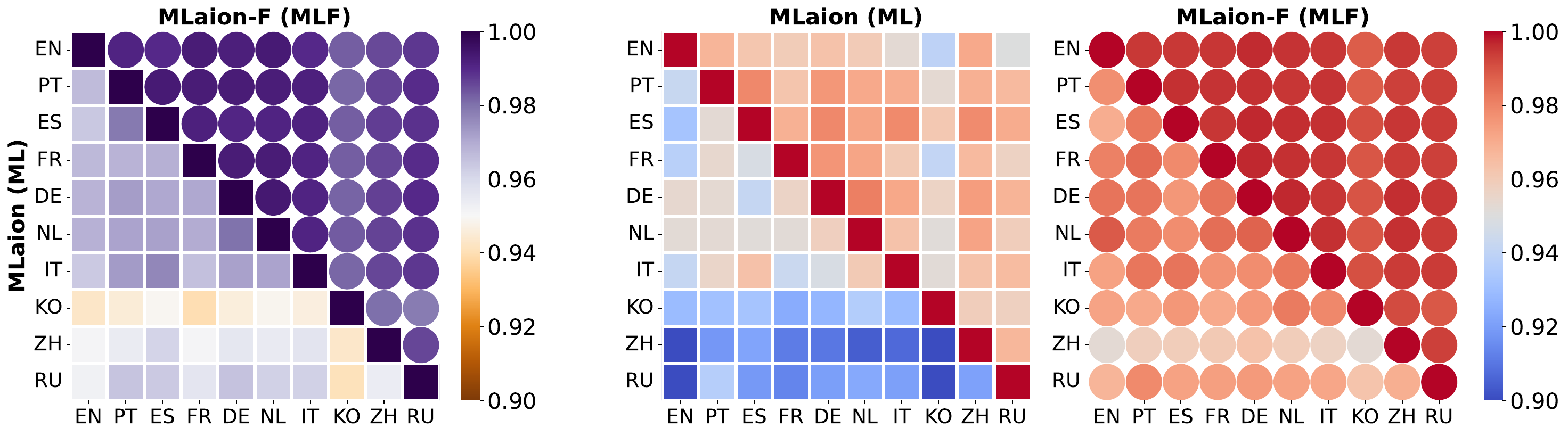}
  \vspace{-0.25cm}
  \caption{Pearson correlation scores between different languages, for the original multilingual CLIPScore model (squared cells) and our finetuned version (circular cells). The first heatmap considers the complete set of instances from the VICR dataset, reporting results for both the original and finetuned model versions (lower/upper diagonal values). The second and third heatmaps consider the subset of instances with COMETKiwi scores below/above the 25th/75th percentile value for each language (lower/upper diagonal values), for the original multilingual CLIPScore model and our finetuned model version, respectively.}
  \vspace{-0.25cm}
  \label{fig:heatmap}
\end{figure*}

Figure~\ref{fig:heatmap} presents Pearson correlation scores between languages, for the best multilingual CLIPScore model (presented in squared cells), and for our finetuned version (presented in circular cells). 
The image contains three heatmaps:
\begin{itemize}[]\setlength\itemsep{-0.3em} 
    \item The first, on the left, shows CLIPScore correlations for the full VICR test dataset, comparing the finetuned model (upper triangle with circular cells) and the pre-finetuned model (lower triangle with squared cells).
    \item The second and third heatmaps display correlations based on translation quality. They focus on translations with COMETKiwi scores in the bottom 25\% (lower triangle) and top 25\% percentile (upper triangle). The second heatmap with squared cells represents the pre-finetuned model, and the third heatmap with circular cells shows our finetuned version.
\end{itemize}

In the left heatmap of Figure~\ref{fig:heatmap}, which includes the entire VICR test set, we observe consistently high correlation values across all languages. The results indicate that CLIPScore correlations are influenced by the quality of the translated captions.

The second and third heatmaps show slightly higher correlations for high-quality translations (upper triangles) compared to the poorest translations (lower triangles). However, even the poorest translations still demonstrate relatively high correlations, suggesting that the overall quality of the translations remains strong across different languages.

It is worth noting that, as expected, the most impacted languages are those that use a different script, particularly in the non-finetuned case. We also see a significant improvement in correlations between languages when using our finetuned CLIPScore model. Although this improvement is expected, given that the finetuned model was trained on in-distribution data, it also reflects the high quality of our multilingual training data, further validating our translation strategy.

\subsubsection{Multilingual Classification} \label{sec: multilingual_classification}

This section explores the robustness of the multilingual CLIPScore assessments through different types of classification tasks. Inspired by previous work~\cite{hessel2021clipscore,sarto2023positive} which assessed accuracy in English-only benchmarks, our goal is to delve deeper into the nuanced realm of multilingual and multicultural understanding.

\paragraph{Robustness to Linguistic Phenomena:}
\begin{table*}[t!]
\centering
\renewcommand{\arraystretch}{1.3}
\resizebox{2\columnwidth}{!}{%
\begin{tabular}{lccccccccccclc}
\multicolumn{11}{c}{Proposed Models}                                                                           &  & \multicolumn{2}{c}{VALSE Models} \\ \cline{1-11} \cline{13-14} 
Finetuning Method   & English  & German  & French  & Spanish  & Chinese  & Portuguese  & Italian  & Russian  & Korean  & Dutch  &  & Model              & English            \\ \cline{1-11} \cline{13-14} 
MLF Both Losses        & \textbf{69.7} & \textbf{67.3} & \textbf{67.5} & \textbf{67.9} & 66.9 & \textbf{69.0} & \textbf{66.0} & \textbf{67.1} & \textbf{60.6} & \textbf{66.2} &  & CLIP               & 64.0           \\
MLF Pearson Loss     & 69.2 & 67.2 & 66.8 & 67.4 & \textbf{67.3} & 67.2 & 66.0 & 65.0 & 59.4 & 65.8 &  & LXMERT             & 59.6           \\
MLF Contrastive Loss & 68.4 & 65.0 & 65.5 & 65.7 & 65.6 & 66.4 & 64.6 & 64.8 & 59.2 & 63.7 &  & ViLBERT            & 46.4            \\
ML        & 67.6 & 64.8 & 64.2 & 65.6 & 64.0 & 65.4 & 64.4 & 63.4 & 58.4 & 62.5 &  & 12-in-1            &  \textbf{75.1} \\\cline{1-11} \cline{13-14}       
\end{tabular}
}
\vspace{-0.25cm}
\caption{Average accuracy scores for the different classification tasks present in the VALSE dataset and its multilingual variants, considering different CLIP models. \textbf{Bold} values signify the best score per language.}
\label{tab:valse-results}
\vspace{-0.25cm}
\end{table*}
Some of the experiments used our machine translated versions of the VALSE dataset, designed to evaluate the robustness to phenomena such as inconsistencies in numeric quantities or spatial relations~\cite{parcalabescu2022valse}. VALSE comprises seven tests that encompass a range of linguistic structures. In each test, a model is presented with a visual input and is tasked with distinguishing true captions from altered versions (i.e., foils), modified to exhibit a specific feature. The evaluation is based on the assumption that the CLIPScore values for true captions should be higher than those for the corresponding foils. Additional information about the dataset is provided in Appendix~\ref{app:datasets}. 

Table~\ref{tab:valse-results} displays the average performance across different language variants. The results show that our finetuned model delivers the highest average performance across nearly all languages. Compared to the models reported in the original VALSE dataset paper, our MultiLingual Finetuned (MLF) model was only surpassed by the multi-task ViLBERT 12-in-1 model~\cite{lu202012}.

Table~\ref{tab:valse-extra} in Appendix \ref{app:results} contains a more detailed breakdown of the performance across the different tests within VALSE. We observe a significant improvement of 6\% to 10\%, on average, in the existence quantifier, plurality, and counting adversarial tests with the finetuned model. For the remaining tests, we saw a more modest performance increase, ranging from 1\% to 5\% compared to the non-finetuned model, with the exception of the action replacement and co-reference tests where performance did not improve. The lower performance for the co-reference tests likely stems from the nature of the task and limitations of our evaluation method. CLIPScore, which was designed to evaluate declarative captions, struggles to effectively score a caption that combines both a question and a yes/no answer. This mismatch may explain the lower results for coreference handling, and can be a key factor for the higher performance of ViLBERT 12-in-1 in the overall average score shown in Table~\ref{tab:valse-results}, when compared to CLIPScore strategies.

In the VALSE experiments, we also observed that different loss functions contributed distinct benefits to various test scenarios. For example, the Pearson loss proved particularly effective in the existence quantifier and action actant swap tests, compared to the contrastive loss. In contrast, the contrastive loss delivered superior performance in the foil-it and the counting adversarial tests. This demonstrates the advantages of our proposed training strategy, which combines both losses to maximize performance.

\begin{table}[t!]
\centering
\renewcommand{\arraystretch}{1.2}
\resizebox{\columnwidth}{!}{%
\begin{tabular}{clccc}
& Model & Accuracy 1 / std & Accuracy 2 / std & Accuracy 3 / std \\ \cline{3-5} \hline
\multirow{3}{*}{\rotatebox{90}{Arabic}} & ML & 84.6 / 1.00 & \textbf{76.3} / 0.63 & \textbf{44.7} / 1.42 \\
& MLF & \textbf{85.4} / 0.92 & \textbf{76.3} / 0.69 & 41.8 / 1.39 \\
& EN & 82.2 / 1.01 & 73.4 / 0.65 & 38.82 / 1.42 \\ \hline
\multirow{3}{*}{\rotatebox{90}{French}} & ML & 86.7 / 0.89 & 77.9 / 0.63 & 45.4 / 1.44 \\
& MLF & \textbf{90.5} / 0.76 & \textbf{79.9} / 0.62 & \textbf{47.7} / 1.46 \\
& EN & 86.2 / 0.83 & 76.9 / 0.66 & 43.75 / 1.44 \\ \hline
\multirow{3}{*}{\rotatebox{90}{Spanish}} & ML & 86.2 / 0.90 & 77.7 / 0.65 & \textbf{46.7} / 1.46 \\
& MLF & \textbf{87.8} / 0.84 & \textbf{78.4} / 0.67 & 45.1 / 1.38 \\
& EN & 87.3 / 0.88 & 77.3 / 0.65 & 44.41 / 1.44 \\ \hline
\multirow{3}{*}{\rotatebox{90}{Russian}} & ML & 87.5 / 0.88 & 78.0 / 0.63 & 44.4 / 1.46 \\
& MLF & \textbf{88.9} / 0.83 & \textbf{79.2} / 0.62 & \textbf{46.7} / 1.48 \\
& EN & 88.6 / 0.84 & 78.2 / 0.64 & 44.08 / 1.51 \\ \hline
\multirow{3}{*}{\rotatebox{90}{Overall}} & ML & 86.3 / 0.92 & 77.8 / 0.64 & 45.1 / 1.45 \\
& MLF & \textbf{88.4} / 0.84 & \textbf{78.8} / 0.65 & \textbf{45.9} / 1.43 \\
& EN & 86.1 / 0.89 & 76.5 / 0.65 & 42.8 / 1.45 \\ \hline
\end{tabular}
}
\vspace{-0.25cm}
\caption{Accuracy for different classification tasks defined over the datasets derived from XVNLI. \textbf{EN:} English-only LAION-H-14 model. \textbf{Bold} values signify the best value per language/task. The std
values in each column represent the standard deviation in the test set for each metric across different models. We used a
stratified random sampling approach, performing 1,000 iterations to generate subsets containing 80\% of the original
data, as we do not finetune all models.}
\label{tab:xvnli}
\vspace{-0.25cm}
\end{table}

\begin{table}[t!]
\centering
\renewcommand{\arraystretch}{1.3}
\resizebox{0.77\columnwidth}{!}{%
\begin{tabular}{clcc}
 & Model & Accuracy 1 / std & Accuracy 2 / std \\ \hline
\multirow{3}{*}{\rotatebox{90}{\small Indonesian}} & ML & 92.4 / 0.35 & 82.3 / 1.13 \\
& MLF & \textbf{93.4} / 0.39 & \textbf{83.3} / 1.14 \\
& EN & 91.4 / 0.39 & 81.2 / 1.18 \\ \hline
\multirow{3}{*}{\rotatebox{90}{\small Mandarin}} & ML & 89.4 / 0.44 & 80.1 / 1.23 \\
& MLF & \textbf{91.5} / 0.49 & \textbf{81.3} / 1.23 \\
& EN & 91.1 / 0.53 & 79.9 / 1.27 \\ \hline
\multirow{3}{*}{\rotatebox{90}{\small Swahili}} & ML & \textbf{85.0} / 0.58 & 65.3 / 1.42 \\
& MLF & \textbf{85.0} / 0.58 & 65.3 / 1.38 \\
& EN & 84.3 / 0.44 & \textbf{70.1} / 1.22 \\ \hline
\multirow{3}{*}{\rotatebox{90}{\small Tamil}} & ML & 85.7 / 0.49 & 74.7 / 1.25 \\
& MLF & 88.0 / 0.51 & 77.5 / 1.28 \\
& EN & \textbf{92.8} / 0.58 & \textbf{85.3} / 1.47 \\ \hline
\multirow{3}{*}{\rotatebox{90}{\small Turkish}} & ML & \textbf{93.5} / 0.37 & \textbf{87.0} / 1.04 \\
& MLF & 92.9 / 0.37 & 86.0 / 0.96 \\
& EN & 87.7 / 0.38 & 79.1 / 0.98 \\ \hline
\multirow{3}{*}{\rotatebox{90}{\small Overall}} & ML & 89.4 / 0.45 & 80.1 / 1.21 \\
& MLF & \textbf{91.5} / 0.47 & \textbf{81.3} / 1.20 \\
& EN & 91.1 / 0.46 & 79.9 / 1.22 \\ \hline
\end{tabular}
}
\vspace{-0.25cm}
\caption{Accuracy for different classification tasks defined over the datasets derived from MaRVL. \textbf{EN:} English-only LAION-H-14 model. \textbf{Bold} values signify the best value per language/task. The std
values in each column represent the standard deviation in the test set for each metric across different models. We used a
stratified random sampling approach, performing 1,000 iterations to generate subsets containing 80\% of the original
data, as we do not finetune all models.}
\label{tab:marvl}
\vspace{-0.25cm}
\end{table}

\paragraph{Classification of Multicultural Instances:}

We also used natively multilingual datasets (i.e., XVNLI and MaRVL) to assess the multilingual and multicultural capabilities of CLIPScore models. 

These datasets originally aimed at assessing semantic inference between the contents of images and texts, featuring positive and negative instances whose challenges for interpretation are more likely to lie on fine-grained semantic understanding and compositional reasoning instead of object detection, and with captions featuring slight variations in how the contents of an image are expressed.

\begin{figure}[h!]
  \centering
  \includegraphics[width=\columnwidth]{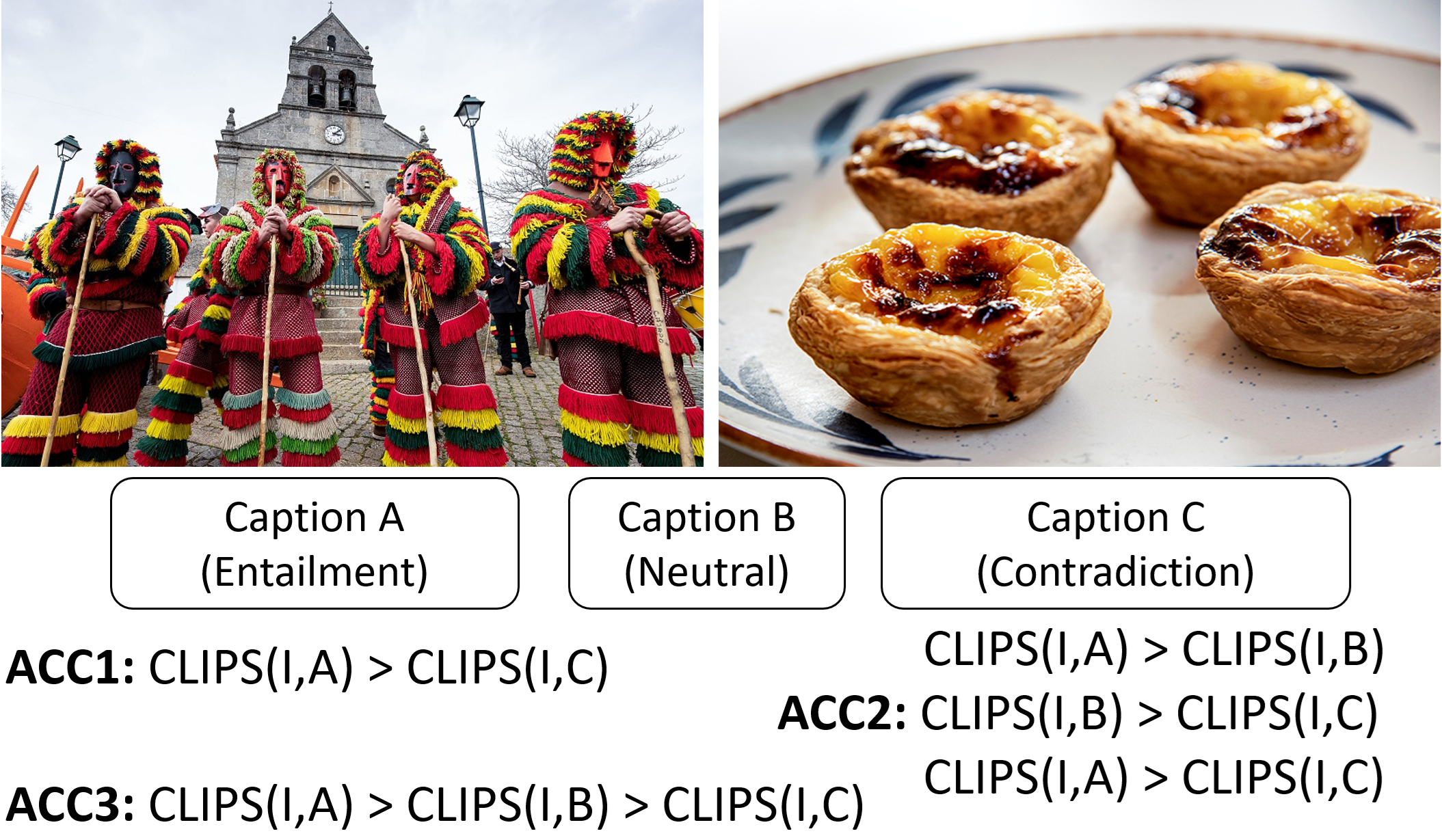}
  \caption{The three different XVNLI multilingual classification tasks, where accuracy is defined with basis on comparisons between CLIPScore values.}
  \label{fig:xvnli-acc}
\end{figure}

Each instance in the XVNLI dataset contains an image-caption pair and a categorical label associated with the relationship between the pair. This label can be either (a) contradiction, (b) neutral, or (c) entailment. Based on these labels, we defined three multilingual classification tasks under this scenario, leveraging concordant/discordant instances as illustrated in Figure~\ref{fig:xvnli-acc}:

\begin{itemize}[align=left]\setlength\itemsep{-0.3em}
    \item [\textbf{Task 1:}] This setting only considers contradiction and entailment instances, under the assumption that the order of the CLIPScore values should match the order of the labels.
    \item [\textbf{Task 2:}] We consider a larger set of duplets and the ordering between the three possible labels. 
    \item [\textbf{Task 3:}]  This setting considers the three possible labels, but we now assess triples of instances $A$, $B$, and $C$, sharing the same image. We only assume a correct classification when we achieve a perfect match between the order of the labels and the CLIPScore values.
\end{itemize}
    



In the case of the MaRVL dataset, each instance contains a caption, two images, and a boolean label with the value "true" when the caption accurately matches both images, and "false" when the caption matches at maximum only one of the images. The data can be analysed considering two or four instances simultaneously, sharing the same caption but featuring distinct pairs of images. We devise two evaluation tasks based on MaRVL.

\begin{itemize}[align=left]\setlength\itemsep{-0.3em} 
    \item [\textbf{Task 1:}] We consider only two instances at a time. We check whether the highest CLIPScore among the images in a true-labeled instance is greater than the lowest CLIPScore among the images in a false-labeled instance. This accounts for the fact that a false instance may contain one image that aligns with the caption.
    \item [\textbf{Task 2:}] We take a more challenging approach by considering four instances sharing the same caption. The objective in this case is for the highest CLIPScores (both of them) from the true-labeled instances to exceed the lowest CLIPScores from the false-labeled instances
\end{itemize}


Our objective in rearranging both datasets into the aforementioned classification tasks (several tests with increasing difficulty) was to allow for the evaluation of the ability of CLIPScore models to reason with fine-grained semantic distinctions in natively multilingual and multicultural contexts.

In this set of multilingual experiments, we assessed three distinct models: the original multilingual LAION/ViT-H-14 (ML), our finetuned variant (MLF), and the English-only LAION/ViT-H-14 (EN). Apart from the multilingual setup, we employed our translation scheme to convert the native non-English data into English, in order to evaluate English models within these native multilingual scenarios. This approach allows us to compare the benefits of leveraging machine translation during training versus inference, and an indirect assessment of the quality of our MT process.

Table~\ref{tab:xvnli} summarizes the results for the classification tasks from XVNLI. As expected, accuracy declined from Task 1 to 3 for all model variants, indicating that model performance tends to decrease as task complexity increases. Additionally, the English-only model performs worse than our MultiLingual Finetuned (MLF) model across all languages and tasks, with particularly noticeable gaps in Arabic and French (languages for which MLF is also outperformed by the ML variant). This suggests that while machine translation can make non-English data accessible to English-only models, it leads to performance degradation, potentially forgoing some cultural nuances. 

Table ~\ref{tab:marvl}, shows a similar trend for MaRVL, i.e., our MLF model performs on par or even better than other variants for most languages, but displays different patterns for low-resource languages like Tamil and Swahili. The English-only model also performs better in the more difficult accuracy task. This behaviour might reflect the challenges multilingual models face in extremely low-resource languages, where the occurrence is scarce to non-existent in the pretraining data. Additionally, it may relate to the imbalanced quality of out-of-English versus into-English MT, with the latter being better for low-resource languages.

Overall, our findings indicate that the MLF model achieved superior results compared to the original ML model, emphasizing the effectiveness of our training strategy. 

In addition, our analysis using natively multilingual datasets (XVNLI and MaRVL) confirms that a translation pipeline with an English-centric model performs worse than directly using a multilingual model. While back-translating captions into English for evaluation with an English-only model has been explored in prior studies ~\cite{wada2023jaspice}, our findings reveal that this approach significantly increases computational costs and response times without offering any practical advantages during inference, even when translation quality is high. Instead, our results highlight the greater value of machine translation for cost-effective dataset expansion during finetuning, which consistently enhances multilingual CLIP model performance.

A more detailed explanation for both the XVNLI and MaRVL experiments is given in Appendix \ref{app:multilingual_experiments}.

While the experiments with the XVNLI and MaRVL datasets provide interesting insights into the effectiveness of multilingual CLIPScore models, they also involve several important limitations. Considering the XVNLI experiments, previous studies have reported good results in multimodal inference leveraging CLIP~\cite{song2022clip}. However, the authors of SNLI-VE~\cite{xie2019visual}, from which XVNLI is derived, noted that good performance (i.e., an accuracy up to 67\%) can be achieved when looking only at the information in the textual hypothesis, without the visual premise, pointing to significant biases in the XVNLI data. In the case of the MaRVL experiments, given that the captions refer to a pair instead of individual images, the CLIPScore values can be unreliable when attempting to match images to textual sentences. Previous studies have noted that CLIP models can treat inputs as a bag-of-words and suffer from a concept association bias~\cite{yamada2022lemons}, e.g. ignoring the missing information when two concepts are present in one of the inputs while the other only contains a single concept. Hence, there is room for future work that pushes even further the analysis of the robustness of caption evaluation metrics. We note that the proposed process to design classification tasks can eventually transfer to newer datasets that address these limitations.

\section{Conclusions}

This study highlights the importance of expanding image captioning evaluation to include multilingual and multicultural research, encouraging more inclusive frameworks in this field. Using a machine translation scheme with quality filtering, we can cost-effectively extend well-established English-centric benchmarks to multiple languages, without compromising benchmark quality and validity, which can be very beneficial for the finetuning and evaluation of new, multilingual evaluation models.

We also propose a finetuning strategy to better leverage and learn from both multi-cultural data and human preferences, and test our models on a set of different datasets and tasks. Our findings show that multilingual models trained with the same amount but with less English-specific data perform equally well on English tasks, while excelling in multilingual and multicultural ones. This reveals the potential of multilingual models to generalize across languages, making them more versatile assets. Additionally, our finetuning approach significantly boosted the model's ability to handle complex linguistic challenges, such as quantifiers, plurality, and numeric inconsistencies, highlighting its adaptability to more intricate language patterns.

Further to machine translation of English data, we also propose a strategy to adapt multilingual datasets from other tasks to support captioning evaluation. The integration of natively multilingual and multicultural datasets into both training and evaluation processes mitigates cultural information loss, reinforcing the reliability of our proposed pipeline for training and evaluating multilingual CLIP models, and making them effective tools for real-world multilingual and multicultural evaluation.

Overall, our work contributes to multilingual captioning evaluation, both in terms of modelling and benchmarking. We hope it will inspire and support further work in this under-researched field.

\newpage
\section*{Limitations and Ethical Considerations}

Although our work does not raise new ethical issues within the domain of vision-language models (e.g., we conducted our experiments on public datasets carefully designed for academic research and extensively used in previous studies), there are still some concerns which we describe below. 

Models like CLIP are, for instance, notorious for their internal biases, e.g. inherited from the training data itself. We, therefore, recommend caution in the use of the approach proposed in this paper and anticipate further research into the specific issue of model biases before relying on our work beyond research environments. Another important limitation in the work reported in this paper concerns the use of machine translated data in some of the evaluation experiments, which, despite our best efforts to avoid translation errors, can still lead to different types of biases and to the reliance on artificially impoverished language. The development of manually curated benchmarks, specifically designed for the assessment of multilingual metrics for image captioning evaluation, is left as an important challenge for future work.

We also note that we used GitHub Copilot\footnote{\url{https://github.com/features/copilot}} during the development of our research work, and we used ChatGPT\footnote{\url{https://openai.com/chatgpt/}} for minor verifications during the preparation of this manuscript.

\section*{Acknowledgements}

We thank the anonymous reviewers for their valuable comments and suggestions. This research was supported by the Portuguese Recovery and Resilience Plan through project C645008882-00000055 (i.e., the Center For Responsible AI), by Fundação para a Ciência e Tecnologia (FCT) through the projects with references 2024.07385.IACDC and UIDB/50021/2020 (DOI:10.54499/UIDB/50021/2020), by EU's Horizon Europe Research and Innovation Actions (UTTER, contract 101070631), and also by FCT/MECI through national funds, and when applicable co-funded EU initiatives, under contract UID/50008 for Instituto de Telecomunicações.

\bibliography{custom}

\appendix
\clearpage
\section{The CLIPScore Metric} \label{app:clipscore}

We now formally describe the CLIPScore and RefCLIPScore metrics~\cite{hessel2021clipscore}, which in our study are assessed in multilingual image captioning scenarios. In brief, CLIPScore is based on a modified cosine similarity between representations for the input image and the caption under evaluation. The image and the caption are both passed through the respective feature extractors from a given CLIP model. Then, we compute the cosine similarity of the resultant embeddings, adjusting the resulting value through a re-scaling operation. For an image with visual CLIP embedding $\textbf{v}$ and a candidate caption with textual CLIP embedding $\textbf{c}$, a re-scaling parameter is set as $w = 2.5$ and we compute the corresponding CLIPScore as follows:
\begin{equation}
\text{CLIPScore}({\textbf{c}}, {\textbf{v}}) =  w \times \max(\cos({\textbf{c}}, {\textbf{v}}), 0).
\end{equation}

To compute a corpus-level CLIPScore, e.g. for evaluating the overall quality of a captioning method over a given dataset of images, we can simply average over all the image-candidate pairs. 

Note that CLIPScore does not depend on the availability of underlying references for each of the images in an evaluation dataset. However, an extension named RefCLIPScore was also proposed, which additionally extracts the vector representations $\textbf{R}$ of each available reference with the CLIP text encoder, and computes the harmonic mean of the CLIPScore value from Equation 3, and the maximal reference cosine similarity:
\begin{align}
\text{RefCLIPScore}&({\textbf{c}}, {\textbf{R}}, {\textbf{v}}) = \notag \\
\text{H-Mean}&( \text{CLIPScore}({\textbf{c}}, {\textbf{v}}), \\
            &\max(\max_{{\textbf{r}} \in {\textbf{R}}} \cos({\textbf{c}}, {\textbf{r}}), 0)).\notag
\end{align}

\section{The Correlation Metrics} \label{app:metrics}
This appendix presents a formal definition of the metrics used in the correlation experiments.

Seeing each of our evaluation datasets as sets of $n$ observations with the form $(\hat{y}_1,y_1), \ldots, (\hat{y}_n,y_n)$, for CLIPScore values $\hat{y}_i$ and reference ratings $y_i$, the Spearman correlation coefficient $\rho$ is defined as the Pearson correlation between the results of converting the scores $\hat{y}_i$ and $y_i$ to ranks. 

Instead of using ranks, we can also define any pair of observations $(\hat{y}_{i},y_{i})$ and $(\hat{y}_{j},y_{j})$, where $i < j$, as concordant (or otherwise discordant) if the sort order of the instances agrees (i.e. if either both $\hat{y}_{i} > \hat{y}_{j}$ and $y_{i} > y_{j}$ holds, or both $\hat{y}_{i} < \hat{y}_{j}$ and $y_{i} < y_{j}$). Based on pairs, the Kendall $\tau$ correlation coefficient assesses the strength of association between the CLIPScore values and the reference ratings, with the $\tau_b$ variant accounting for ties and being defined as:
\begin{equation}
\tau _{B}={\frac {n_{c}-n_{d}}{\sqrt {(n_{0}-n_{1})(n_{0}-n_{2})}}},
\end{equation}
where $n_{c}$ is the number of concordant pairs, $n_{d}$ the number of discordant pairs, $n_{0} = n(n-1)/2$, $n_{1} = \sum _{i}t_{i}(t_{i}-1)/2$, $n_{2} = \sum _{j}u_{j}(u_{j}-1)/2$, $t_{i}$ is the number of tied values in the $i^\text{th}$ group of ties for the CLIPScore, and  $u_{j}$ is the number of tied values in the $j^\text{th}$ group of ties for the reference ratings.

In turn, $\tau_c$ accounts with the fact that the underlying scales of the scores are different for CLIPScore and the reference ratings, being defined as:
\begin{equation}
\tau_{c}={\frac {n_{c}-n_{d}}{n_{0}}} \times {\frac {n-1}{n}} \times {\frac {m}{m-1}},
\end{equation}
where $m$ is the number of values in the ranking scale for the reference ratings.

\section{Description of the Datasets}\label{app:datasets}

The following datasets were used in the tests that assessed correlation with human judgment.
\begin{itemize}\setlength\itemsep{0em}
\item \textbf{Flickr8K-Expert \cite{hodosh2013framing}:}
This dataset comprises $16,992$ expert human judgments for $5,664$ image-caption pairs from the Flickr8K dataset. Human assessors graded captions on a scale of 1 to 4, where 4 indicates a caption that accurately describes the image without errors, and 1 signifies a caption unrelated to the image.

\item \textbf{Flickr8K-CF \cite{hodosh2013framing}:} This dataset consists of $145,000$ binary quality judgments collected with CrowdFlower, involving  $47,830$ image-caption pairs with $1,000$ unique Flickr8K images. Each pair received at least three binary judgments, and we use the proportion of {\it yes} annotations as the score for each pair.

\item \textbf{Composite \cite{aditya2015images}:} This dataset contains $13,146$ image-caption pairs taken from MSCOCO (2007 images), Flickr8K (997 images), and Flickr30K (991 images). Each image originally had five reference captions. One of these references was chosen for human rating and subsequently removed from the reference set that is to be used when assessing evaluation metrics.

\item \textbf{VICR \cite{narins2024validated}:} The Validated Image Caption Rating (VICR) dataset features 68,217 ratings, collected through a gamified approach, for 15,646 image-caption pairs involving 9,990 distinct images. The authors of the dataset demonstrated that it exhibits a superior inter-rater agreement compared to other alternatives (e.g., an improvement of 19\% in Fleiss’ $\kappa$ when compared to the agreement for the Flickr8K-Expert dataset), and it features a more balanced distribution across various levels of caption quality. In our tests, we used the test split of the VICR dataset, which includes 3,161 image-caption pairs, with 2,000 images from the MSCOCO 2014 validation dataset and 1,161 images from the Flickr8K dataset. When using VICR to finetune CLIP models with a contrastive loss, we used the original image captions from MSCOCO or Flickr8K.
\end{itemize}

All the previous datasets are originally available only for English, but we translated them to nine other different languages using the approach described in Appendix~\ref{app:translation}.

For the experiments that assessed accuracy in terms of distinguishing correct vs incorrect captions, we used the following datasets.
\begin{itemize}\setlength\itemsep{0em}
\item \textbf{VALSE \cite{parcalabescu2022valse}}: VALSE is designed to test visio-linguistic grounding capabilities on specific linguistic phenomena. It is composed by seven tasks, each with the same structure: given a visual input, a model is asked to distinguish real captions from foils, where a foil is constructed from a caption by altering a word or phrase that realizes a specific linguistic phenomenon. The tests include: (a) existential quantifiers, where models need to differentiate between examples (i) where there is no entity of a certain type or (ii) where one or more of these entities are visible in an image; (b) plurality, where models need to distinguish between noun phrases denoting a single entity in an image ({\it exactly one flower}), versus multiple entities ({\it some flowers}); (c) counting, where models needs to differentiate between examples where the specific number of entities in the associated image is correct or incorrect, given the statement; (d) spatial relations, where models need to distinguish between different spatial relations, with foils differing from the original caption only by the replacement of a spatial preposition; (e) actions, particularly (i) action replacement and (ii) actant swaping, where models need to (i) identify whether an action mentioned in the text matches the action seen in the image (e.g., {\it a man shouts} versus {\it smiles at a woman}), and (ii) correctly identify the participants of an action and the roles they play (e.g., is it the man who is shouting or is it the woman); (f) coreference, where models need to perform pronominal coreference resolution, encompassing cases where (i) the pronoun has a noun (phrase) antecedent and pronoun and (noun) phrase are both grounded in the visual modality (e.g., in {\it a woman is driving a motorcycle}, is she wearing a helmet?), and cases where (ii) the pronoun refers to a region in the image or even to the entire image (e.g., {\it is this outside?}); (g) foil-it cases, in which the foil minimally differs from the original caption, only by swapping a important noun.

\item \textbf{XVNLI \cite{bugliarello2022iglue}: } XVNLI is a multilingual dataset for evaluating vision-language inference, challenging models to predict entailment relationships between a textual hypothesis and an image premise. XVNLI includes high/mid-resource languages like Arabic, French, Spanish, Russian, and English. This dataset includes 1,164 instances per language, each featuring an image and two captions in different languages. There are 357 unique images in total.

\item \textbf{MaRVL \cite{liu2021visually}: } MaRVL follows a format similar to the English NLVR2 dataset~\cite{suhr2019corpus} and is designed as a multicultural vision-language reasoning dataset, where the goal is to determine the correctness of a sentence about a pair of images. MaRVL predominantly comprises very low-resource languages: Indonesian, Chinese, Swahili, Tamil, and Turkish. This dataset includes around one thousand instances per language, each featuring two image and one caption. There are 1,411 unique captions in total. The English content is composed by collecting the reverse translations from the low-resource languages into English, as provided by the authors on the original GitHub repository\footnote{\url{https://github.com/marvl-challenge/marvl-code/tree/master/data}}. 
\end{itemize}

We also used an additional English dataset in experiments reported in Appendix \ref{app:english_experiments}.
\begin{itemize}
    \item {\textbf{Pascal-50S} \cite{vedantam2015cider}}: The dataset features preference judgments between pairs of sentences associated to images. There are a total of 4K sentence pairs, evenly split across four categories, such as two human captions, two machine captions, and so on. For each pair, 48 human pairwise judgments were collected\footnote{Instead of being presented with the image, the human annotators were presented only with a reference caption (and the two caption candidates to rank).}. Following prior work, instead of computing correlation coefficients, accuracy is computed. Specifically, the caption preferred by a majority of annotators to be correct is considered, and we measure how often the evaluation metric assigns a higher score to the preferred caption in the pair. Ties are broken randomly.
\end{itemize}

For model training, besides instances in the training split from the aforementioned VICR dataset, we also used data from the natively multilingual CrossModal-3600 dataset (i.e., XM3600, in short). 

\begin{itemize}
\item \textbf{XM3600~\cite{thapliyal2022crossmodal}:} This is a geographically-diverse set of 3600 images annotated with human-generated reference captions in 36 languages. The images were selected from all across the world, covering regions where the 36 languages are spoken, and consistently annotating captions in terms of style across all languages, while avoiding annotation artifacts due to direct translation.
\end{itemize}

\section{The Machine Translation Scheme} \label{app:translation}
This appendix describes the translation scheme that was used to machine translate the datasets used in our experiments. This scheme is designed to mitigate low-quality translations, or hallucinations generated by the machine translation model, thus providing reliable datasets at the end. We specifically used a large (i.e., 1.2 billion parameters) open-access multilingual machine translation model named M2M100~\cite{m2m100}, available on the HuggingFace\footnote{ \url{https://huggingface.co/facebook/m2m100_1.2B}}  model hub. M2M100 was trained on a range of high and low-resource languages from different families and using different scripts, achieving state-of-the-art performance across a diverse set of 100 languages. 

While machine translated data allows us to assess multilingual captioning metrics, the results will depend not only on the performance of the metrics but also on the quality of the translations. Low-quality translations, or hallucinations generated by the translation model, will impact the caption and break our assumption that human ratings for the English data can be transferred across languages. To address this issue, we propose to use the COMETKiwi~\cite{rei2022cometkiwi,rei2023scaling} machine translation quality estimation metric to control for translation quality, assessing the impact of low quality translations on the observed results. 

We specifically began by translating the VICR dataset, followed by the other English datasets mentioned in Appendix D. VICR features English captions with human ratings and also reference captions originally from the MSCOCO and Flickr8K datasets. For each caption, whether a candidate or a reference, we return 25 translations using a beam search technique with 100 beams. Subsequently, we filtered the candidates with a language checker, to ensure proper translation into the intended language. After the language check, we selected for each instance the translation that scored higher based on a large COMETKiwi model\footnote{\url{https://huggingface.co/Unbabel/wmt23-cometkiwi-da-xxl}}.

\section{Assessments on English Data} \label{app:english_experiments}

\begin{table*}[t!]
\centering
\renewcommand{\arraystretch}{1}
\resizebox{1.95\columnwidth}{!}{%
\begin{tabular}{c l c c c c c c c c c c c c c c c c}
 & & \multicolumn{3}{c}{VICR} & ~~ & \multicolumn{3}{c}{Expert} & ~~ & \multicolumn{3}{c}{CrowdFlower} & ~~ & \multicolumn{3}{c}{Composite} \\ 
 \cmidrule{3-5} \cmidrule{7-9} \cmidrule{11-13} \cmidrule{15-17} 
 & & $\tau_b$ & $\tau_c$ & $\rho$ & & $\tau_b$ & $\tau_c$ & $\rho$ & & $\tau_b$ & $\tau_c$ & $\rho$ & & $\tau_b$ & $\tau_c$ & $\rho$ \\ \toprule
 \multicolumn{1}{l}{\STAB{\multirow{13}{*}{\rotatebox[origin=c]{90}{Related Work}}}} & 
BLEU1                                      & 57.9 & 63.7 & 74.0 & & 32.2 & 32.3 & 40.4 & & 17.9 & 9.3 & 22.3 & & 45.8 & 46.2 & 63.0 & \\
& BLEU4                                    & 54.8 & 60.4 & 70.5 & & 30.6 & 30.8 & 38.7 & & 16.9 & 8.7 & 21.0 & & 46.4 & 46.9 & 63.7 & \\
& CIDEr                                    & 63.1 & 69.8 & 79.3 & & 43.6 & 43.9 & 54.3 & & 24.6 & 12.0 & 29.3 & & 48.1 & 48.8 & 65.0 & \\ \cmidrule{2-17}
& CLIPScore                                & 60.7 & 67.1 & 76.9 & & 51.1 & 51.5 & 63.1 & & 34.4 & 17.8 & 42.4 & & 50.6 & 51.3 & 67.9 & \\
& RefCLIPScore                             & 66.3 & 73.3 & 82.2 & & 52.0 & 52.4 & 63.7 & & 36.4 & 18.8 & 44.7 & & 56.8 & 57.6 & 74.7 & \\
& CLIP+CIDEr                               & 66.8 & 73.8 & 82.6 & & 53.1 & 53.4 & 65.3 & & 33.9 & 17.5 & 41.8 & & 54.3 & 55.1 & 72.2 & \\ \cmidrule{2-17}
& MID~\cite{kim2022mutual}                 & --   & --   & --   & & --   & 54.9 & --   & & 37.3 & -- & -- & & -- & -- & -- & \\ \cmidrule{2-17}
& InfoMetIC~\cite{hu2023infometic}         & --   & --   & --   & & -- &   54.2 & --   & & 36.3 & -- & -- & & -- & 59.2 & -- & \\
& InfoMetIC$^+$~\cite{hu2023infometic}     & --   & --   & --   & & -- &   55.5 & --   & & 36.6 & -- & -- & & -- & 59.3 & -- & \\ \cmidrule{2-17}
& PR-MCS~\cite{kim2023pr}                  & --   & --   & --   & & -- & 50.6 & 65.6 & & --   & --   & --   & & --   & --   & --   & \\ \cmidrule{2-17}
& PACScore~\cite{sarto2023positive}        & --   & --   & --   & & 53.9 & 54.3 & -- & & 36.0 & 18.6 & -- & & 51.5 & 55.7 & -- & \\
& PACScore++~\cite{sarto2024positive}      & --   & --   & --   & & 54.1 & 54.5 & -- & & 37.0 & 19.1 & -- & & 53.9 & 58.3 & -- & \\
& RefPACScore~\cite{sarto2023positive}     & --   & --   & --   & & 55.5 & 55.9 & -- & & 37.6 & 19.5 & -- & & 53.0 & 57.3 & -- & \\ 
& RefPACScore++~\cite{sarto2024positive}   & --   & --   & --   & & 55.3 & 55.7 & -- & & 37.9 & 19.6 & -- & & 54.7 & 59.1 & -- & \\ \cmidrule{2-17}
& CLAIR (GPT3.5)~\cite{chan2023clair}      & --   & --   & --   & & 61.6 & -- & -- & & -- & -- & -- & & \textbf{60.4} & -- & -- & \\
& CLAIR$_{E}$~\cite{chan2023clair}         & --   & --   & --   & & \textbf{62.7} & -- & -- & & -- & -- & -- & & 59.2 & -- & -- & \\\cmidrule{2-17}
& BRIDGE ViT-B/32~\cite{sarto2024bridge}   & --   & --   & --   & & 54.4 & 54.8 & 66.7 & & 36.1 & 18.7 & 44.5 & & 50.9 & 55.0 & 65.4 & \\
& BRIDGE ViT-L/14~\cite{sarto2024bridge}   & --   & --   & --   & & 55.4 & 55.8 & 67.7 & & 36.3 & 19.0 & 44.7 & & 52.9 & 57.2 & 67.8 & \\\cmidrule{2-17}
& FLEUR~\cite{lee2024fleur}                           & --   & --   & --   & & --   & 53.0 & --   & & 38.6 & --   & -- & & -- & 63.5 & -- & \\
& RefFLEUR~\cite{lee2024fleur}                        & --   & --   & --   & & --   & 51.9 & --   & & \textbf{38.8} & --   & -- & & -- & \textbf{64.2} & -- & \\\cmidrule{2-17}
& VICR~\cite{narins2024validated}          & --   & 75.8 & --   & & -- & 53.1 & --   & & --    & --   & --   & & --   & --  & --   & \\ \cmidrule{2-17}
& Polos~\cite{wada2024polos}               & --   & --   & --   & & --   & 56.4 & -- & & 37.8 & -- & -- & & -- & 57.6 & -- & \\\cmidrule{2-17}
& G-VEval~\cite{tong2024g}                 & --   & --   & --   & & 61.5 & \textbf{59.7} & -- & & 38.7 & \textbf{20.2} & -- & & -- & -- & -- & \\
\cmidrule{2-17}
& CAMScore~\cite{cui2025evaluating}                        & --   & --   & --   & & 54.8 & 55.6 & -- & & 37.5 & 19.3 & -- & & 53.4 & 57.5 & -- & \\

\midrule
\multicolumn{1}{l}{\STAB{\multirow{9}{*}{\rotatebox[origin=c]{90}{CLIPScore variants}}}} 
& English CLIPScore                             & 67.4 & 74.4 & 83.1 & & 56.3 & 56.6 & 68.4 & & 38.5 & 19.9 & 47.1 & & 55.1 & 55.9 & 72.8 & \\
& English RefCLIPScore                          & 68.3 & 75.4 & 83.8 & & 56.8 & 57.1 & 68.9 & & 38.6 & 19.9 & \textbf{47.3} & & 56.4 & 57.2 & 74.0 & \\
& English CLIP+CIDEr~~~~~~~~~~~~~~~~            & 67.6 & 74.8 & 83.3 & & 56.0 & 56.4 & 68.5 & & 35.6 & 18.4 & 43.8 & & 53.9 & 54.7 & 71.7 & \\ \cmidrule{2-17} 
& ML CLIPScore                              & 67.6 & 73.0 & 82.4 & & 57.4 & 54.3 & 67.3 & & 38.2 & 19.4 & 46.2 & & 55.5 & 56.2 & 73.2 & \\
& ML RefCLIPScore                           & 69.1 & 74.6 & 83.6 & & 58.1 & 55.0 & 67.9 & & \textbf{38.8} & 19.7 & 46.9 & & 57.3 & 58.1 & \textbf{75.0} & \\
& ML CLIP+CIDEr~~~~~~~~                     & 67.6 & 74.7 & 83.4 & & 55.5 & 55.9 & 67.8 & & 36.3 & 18.8 & 44.7 & & 55.6 & 56.4 & 73.5 & \\ \cmidrule{2-17} 
& MLF CLIPScore                     & 68.7 & 75.4 & 84.1 & & 59.8 & 57.1 & 70.2 & & 37.8 & 19.3 & 45.9 & & 47.4 & 48.0 & 64.7 & \\
& MLF RefCLIPScore                  & \textbf{69.7} & \textbf{76.5} & \textbf{84.8} & & 60.2 & 57.5 & \textbf{70.6} & & 37.6 & 19.1 & 45.6 & & 53.3 & 54.0 & 70.8 & \\
& MLF CLIP+CIDEr~~~~~~~~            & 68.5 & 75.8 & 84.2 & & 57.3 & 57.7 & 70.0 & & 35.8 & 18.5 & 44.0 & & 52.5 & 53.2 & 70.0 & \\ \bottomrule 

\end{tabular}
}
\vspace{-0.25cm}
\caption{Comparison between published results and our best English, multilingual, and finetuned (also multilingual) CLIPScore models, considering settings (a) without human references, (b) using human references, and (c) combining CIDEr with CLIPScore when using human references. }
\label{tab:old_publish}
\vspace{-0.25cm}
\end{table*} 

Table~\ref{tab:old_publish} compares recent studies, proposing other metrics, against the best performing English-only and multilingual CLIPScore models\footnote{Note that we computed all the correlation scores for the CLIPScore variants and traditional image caption metrics based on lexical matches, whereas the results for other more recent metrics are taken from the corresponding publications.}. We considered different evaluation settings, assessing results (a) without references, (b) using references, and (c) combining CIDER with RefCLIPScore when using references~\cite{qiu2023gender}. 

Results confirm that the original CLIPScore outperforms standard captioning metrics in all evaluated datasets, such as BLEU or CIDEr. The correlations consistently improve with RefCLIPScore, but the combination of RefCLIPScore with CIDEr only improved the smaller CLIP model (by previously published results from \citet{qiu2023gender}), instead decreasing the performance in the cases involving the other CLIPScore variants.  

Our study demonstrates that more powerful CLIP models can compete with, and even surpass, complex and specialized systems in image captioning evaluation. The multilingual model performs competitively with the best English-only models across nearly all metrics, whether human references are used or not. This strong performance in both reference-free and reference-aided settings highlights the potential of multilingual CLIPScore. Notably, our finetuned model outperformed the original multilingual LAION ViT-H/14 model and Apple's best model across all CLIPScore variants on the VICR and Expert datasets, even without references. When references were incorporated, our model achieved the best performance on these datasets, surpassing specialized architectures such as VICR, Polos, InfoMetIC, or RefPACScore. By comparing results against current state-of-the-art systems, we found that our finetuned model delivers competitive results with more complex and specialized image captioning evaluation methods like CAMScore, G-VEval, or CLAIR. 

\begin{table}[t!]
\renewcommand{\arraystretch}{1.4}
\centering
\resizebox{1.00\columnwidth}{!}{%
\begin{tabular}{lcccc}
Results without References       & HC   & HI   & HM   & MM   \\ \hline
CLIP-B/32               & 56.4 & 99.1 & 96.9 & 73.8  \\
PAC-S                   & 60.6 & 99.3 & 96.9 & 72.9  \\
PAC-S++                 & 59.5 & 99.6 & 96.5 & 73.6  \\
BRIDGE-B/32             & 59.4 & 99.4 & 97.5 & 74.0 \\
BRIDGE-L/14             & 61.2 & 99.6 & 96.6 & 74.1  \\
FLEUR                   & 61.3 & \textbf{99.7} & 97.6 & 74.2  \\
CAMScore                & \textbf{68.8}  & 99.6 & 97.4 & 77.4  \\
Best English Model      & 57.0 & \textbf{99.7} & 96.3 & 74.9   \\
ML                      & 57.8 & \textbf{99.7} & 96.2 & 73.6 \\  \hline
MLF                     & 66.2 & 99.6 & 96.0 & \textbf{78.8} \\ 
MLF$_C$                 & 65.8 & 99.6 & 97.0 & 76.7 \\ 
MLF$_P$                 & 65.7 & 99.6 & \textbf{99.3} & 75.1 \\ \hline
\end{tabular}
}
\vspace{-0.25cm}
\caption{Pascal50S accuracy results, comparing different methods against our CLIPScore variants.}
\label{tab:clip-pascal}
\vspace{-0.25cm}
\end{table}
\begin{table}[t!]
\renewcommand{\arraystretch}{1.4}
\centering
\resizebox{1.00\columnwidth}{!}{%
\begin{tabular}{lcccc}
Results with References    & HC   & HI   & HM   & MM   \\ \hline
CLIP-32-B               & 64.8 & 99.7 & 96.4 & 73.9 \\
PAC-S                   & 67.7 & 99.6 & 96.0 & 75.6  \\
PAC-S++                 & 67.2 & 99.6 & 96.2 & 74.2  \\
CLAIR (GPT3.5)          & 52.4 & 99.5 & 89.8 & 78.7  \\
CLAIR$_E$                 & 57.7 & \textbf{99.8} & 94.6 & \textbf{81.9}  \\
FLEUR                   & 68.0 & \textbf{99.8} & \textbf{98.0} & 76.1  \\
Polos                   & 70.0 & 99.6 & 97.4 & 79.0  \\
Best English Model      & 68.1 & \textbf{99.8} & 97.6 & 78.0   \\
ML                      & 64.9 & \textbf{99.8} & 97.6 & 75.2 \\ \hline
MLF                     & \textbf{72.6} & 99.7 & 97.6 & 80.6 \\ 
MLF$_C$                 & 71.7 & 99.7 & \textbf{98.0} & 78.9 \\ 
MLF$_P$                 & 70.5 & 99.7 & 97.0 & 78.5 \\ \hline
\end{tabular}
}
\vspace{-0.25cm}
\caption{Pascal50S accuracy results, comparing different methods against our RefCLIPScore variants.}
\label{tab:refclip-pascal}
\vspace{-0.25cm}
\end{table}

Besides the English datasets with numeric ratings discussed in Appendix \ref{app:datasets}, through which we assessed correlations, we used the Pascal-50S dataset \cite{vedantam2015cider} to evaluate model performance on English data. In this dataset, raters made pairwise judgments between pairs of sentences associated with an image. There are 4K sentence pairs in total, split evenly across four categories:
\begin{itemize}\setlength\itemsep{-0.3em}
    \item HC: Two correct captions written by a human; 
    \item HI: Two captions written by a human, but one of the captions is wrong; 
    \item HM: Two correct captions, with one written by a human and one by an algorithm; 
    \item MM: Two correct captions by an algorithm.
\end{itemize}

For each pair, 48 human pairwise judgments of preferred captions were gathered. Following prior work, instead of computing correlation coefficients, we compute accuracy.

Tables ~\ref{tab:clip-pascal} and ~\ref{tab:refclip-pascal} compare the top-performing English and multilingual models, including our finetuned version of the best multilingual model, on the Pascal-50S benchmark using CLIPScore and RefCLIPScore, respectfully. The results show significant improvements after applying our finetuning strategy, particularly in the most challenging tasks (i.e., HC and MM, where both captions are correct and written by a similar entity). Ablation experiments using model checkpoints, trained solely with the contrastive loss or the Pearson loss (i.e., finetuned model with the contrastive loss (C) and finetuned model with the Pearson correlation loss (P), respectively), demonstrated improvements over the original model, suggesting that both loss functions contribute to enhanced performance.

\section{Multicultural Experiments} \label{app:multilingual_experiments}

This appendix details the datasets and experimental settings that were considered for the tests including natively multilingual and multicultural data.
\subsection{Settings for the XVNLI Experiments}
Each instance in the XVNLI dataset contains an image-caption pair and a categorical label associated with the relationship between the pair. This label can be either (a) contradiction, (b) neutral, or (c) entailment. With basis on the labels, we defined three multilingual classification experiments under this scenario, leveraging concordant/discordant instances as illustrated in Figure~\ref{fig:xvnli-acc}:

\textbf{Experiment 1:} This setting only considers instances with the extreme label classes (i.e., contradiction and entailment), noting that some previous studies have pointed to the fact that SNLI-VE, from which XVNLI is derived, has some problems in the annotations for the neutral class~\cite{kayser2021vil}. We compare pairs of instances $A$ and $B$ with the same image, in which the label associated with $A$ differs from the label associated with $B$. When computing CLIPScore values individually for the instances $A$ and $B$, the order of the CLIPScore values should match the order of the labels (i.e., contradiction $<$ entailment).

\textbf{Experiment 2:} In a more challenging scenario, we can consider a larger set of instances and the ordering between the three possible labels (i.e. entailment $>$ neutral $>$  contradiction), i.e. including also the neutral class. Similarly to the previous case, by fixing an image and comparing pairs of captions associated with that image with different labels, we assess the matching of the order between the labels against the CLIPScore values.

\textbf{Experiment 3:} In this case, we also consider the three possible labels, but we now assess triples of instances $A$, $B$, and $C$ from the dataset, sharing the same image. We only assume a correct classification when we achieve a perfect match between the order of the labels and the CLIPScore values.

\begin{figure*}[t!]
  \centering
  \includegraphics[width=2\columnwidth]{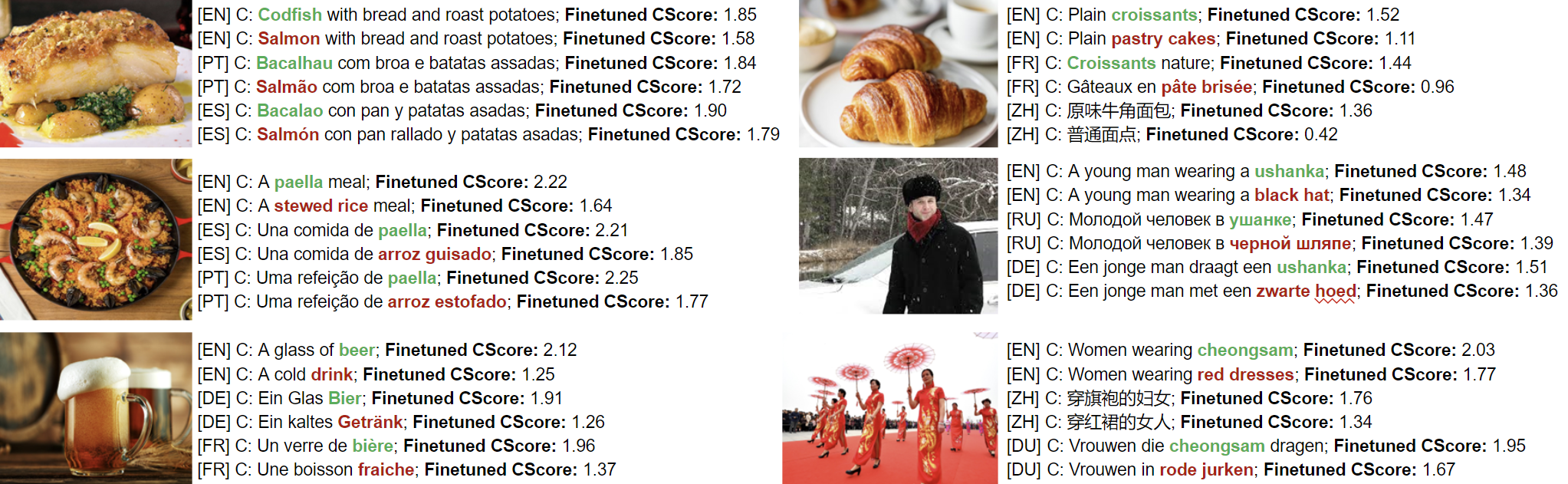}
  \caption{Multilingual CLIPScore values for image-caption pairs featuring concepts biased to particular languages.}
  \label{fig:multicultural}
\end{figure*}

\subsection{Settings for the MaRVL Experiments}
For the MaRVL dataset, each instance consists of a caption, two images, and a boolean label with the value "true" when the caption accurately matches both images, and "false" when the caption is incorrect (i.e., because its contents only describe at maximum one of the images instead of both). The data can be analysed considering two or four instances at a time, sharing the same caption but featuring distinct pairs of images. The MaRVL test was designed in such a way that, within the four instances associated to the same captions, two of them are labeled as "true" while the remaining two are labeled as "false". We consider two multilingual classification experiments under this scenario, defined as follows:

\textbf{Experiment 1:} We draw comparisons between pairs of instances with distinct labels but featuring the same caption. For the instances labeled as "true", we compute the CLIPScore values for the images associated with the caption and select the maximum, obtaining the score for the image that best aligns with the caption. Conversely, we perform a similar computation for the instances labeled as "false", this time choosing the minimum CLIPScore value, which results in the score for the image that least aligns with the caption, presumably the incorrect image. The maximum CLIPScore value in an instance labeled as "true" should be higher than the minimum CLIPScore value of an instance labeled as "false".

\textbf{Experiment 2:} In a more challenging scenario, we consider sets of four instances sharing the same caption (i.e., two instances labeled "false" and two instances labeled "true") and decide on a correct classification if both maximum CLIPScore values of the "true" instances are higher than both the minimum CLIPScore values of the "false" instances. 

\section{A Qualitative Study with Captions Featuring Culturally Related Concepts}
\label{app:qualitative}

We performed a small qualitative study on image-caption pairs that feature concepts where some languages should exhibit a particular bias (e.g., {\it codfish} in the case of Portugal, {\it paella} for Spain, {\it beer} for Germany, {\it croissant} for France, {\it ushanka} for Russia, and {\it cheongsam} for China). We attempted to see if the multilingual CLIPScore could distinguish between two plausible captions, where one mentions a specific concept that should better match the image. Figure \ref{fig:multicultural} shows that the multilingual CLIPScore is indeed capable of distinguishing nuanced multicultural concepts, favouring culturally specific captions over generic ones.

\section{Additional Classification Results} \label{app:results}

Table~\ref{tab:valse-extra} presents classification results on the different tasks from the VALSE dataset, separately for each of the considered languages and comparing different finetuning strategies (i.e., without model finetuning, considering only the contrastive loss, only the Pearson correlation loss, or the combined loss function). Results are, in general, better when considering the combined loss function.

In turn, Table~\ref{tab:multilingual-results-2} presents correlation results on the VICR dataset separately for each language and comparing the same two CLIP models under the different finetuning strategies. Results again show that the smaller CLIP model approaches the performance of the larger model, with better results consistently obtained when considering the combined loss function.

\begin{table*}[t!]
\fontsize{40pt}{40pt}\selectfont
\centering
\resizebox{2\columnwidth}{!}{%
\begin{tabular}{llllcccccccccccccccccccc}
\hline & \\
\multicolumn{3}{c}{} & \multirow{2}{*}{\textbf{Finetuning}} & ~ & \multicolumn{1}{c}{\textbf{Existence}} & ~ & \multicolumn{1}{c}{\textbf{Plurality}} & ~ & \multicolumn{3}{c}{\textbf{Counting}}      & ~ & \multicolumn{1}{c}{\textbf{Sp.Rel}} & ~ & \multicolumn{2}{c}{\textbf{Action}} & ~ & \multicolumn{2}{c}{\textbf{Coreference}}                    & ~ & \multicolumn{1}{c}{\multirow{2}{*}{\textbf{Foil-it!}}} & ~ & \multirow{2}{*}{\textbf{Avg.}} \\ \cline{10-12} \cline{16-17} \cline{19-20}

& & & & & \multicolumn{1}{c}{quantifiers} & ~ & \multicolumn{1}{c}{number} & ~ & balanced & sns. & \multicolumn{1}{c}{adv.} & ~ & \multicolumn{1}{c}{relations} & ~ & repl. & \multicolumn{1}{c}{actant swap} & ~ & standard & \multicolumn{1}{c}{clean} & \multicolumn{1}{c}{}                      &&             &                                \\& \\ \hline & \\
\multicolumn{1}{l}{\STAB{\multirow{4}{*}{\rotatebox[origin=c]{90}{English}}}}
& \multicolumn{1}{l}{}       
& \multicolumn{1}{l}{}                                & Both          & & \textbf{81.6} & & \textbf{76.3} & & \textbf{68.3} & \textbf{75.8} & 66.1 && \textbf{66.5} && 86.0 & \textbf{62.1} & & 46.9 & 42.3 & & 95.0 & & \textbf{69.7} \\
& & \multicolumn{1}{l}{}                               & Pearson      & & 80.8 & & 75.9 & & 67.1 & 74.7 & 65.8 && 64.7 && \textbf{87.5} & 62.0 & & 46.9 & 41.3 & & 94.8 & & 69.2 \\
& & \multicolumn{1}{l}{}                                & Contrastive & & 75.4 & & 75.3 & & 66.7 & 74.4 & \textbf{66.9} && 61.5 & & 87.0 & 58.5 & & 47.6 & 44.2 & & \textbf{95.2} & & 68.4 \\
& & \multicolumn{1}{l}{}                                & None & & 74.7 && 69.6 && 66.9 & 72.4 & 65.3 && 61.7 && 86.7 & 56.1 && 51.3 & 45.2 && 94.2 && 67.6  \\ \\\hline \\      
\multicolumn{1}{l}{\STAB{\multirow{4}{*}{\rotatebox[origin=c]{90}{German}}}}
& \multicolumn{1}{l}{}        
& \multicolumn{1}{l}{}                                & Both && 78.8 && \textbf{71.7} && 68.4 & \textbf{74.6} & 58.5 && \textbf{57.2} && \textbf{77.6} & 62.1 && 53.0 & 43.3 && \textbf{94.9} && \textbf{67.3} \\
& & \multicolumn{1}{l}{}                                & Pearson && \textbf{79.8} && 71.1 && \textbf{68.9} & 73.7 & 55.3 && 56.8 && 77.2 & \textbf{62.5}& & \textbf{54.7} & 45.2 && 94.2 && 67.2 \\
& & \multicolumn{1}{l}{}                                & Contrastive && 64.8 && 71.0 && 66.2 & 72.7 & 61.9 && 54.6 && 76.1 & 59.2 && 48.9 & 45.2 && 94.3 && 65.0 \\
& & \multicolumn{1}{l}{}                                & None & &69.3 && 69.2 && 68.8 & 72.6 & 56.6 && 53.3 && 75.9 & 57.4 && 53.4 & 44.2 && 91.9 && 64.8  \\ \\ \hline \\                               
\multicolumn{1}{l}{\STAB{\multirow{4}{*}{\rotatebox[origin=c]{90}{French}}}}
& \multicolumn{1}{l}{}      
& \multicolumn{1}{l}{}                                & Both && 80.2 && \textbf{75.1} && \textbf{69.7} & 72.4 & 61.4 && \textbf{56.1} && 75.5 & \textbf{61.4} && 48.6 & 48.1 && \textbf{94.5} && \textbf{67.5} \\
& & \multicolumn{1}{l}{}                                & Pearson && \textbf{82.2} && 73.4 && 69.0 & \textbf{72.8} & 56.9 && 55.9 && 75.8 & 61.3 && 46.6 & 47.1 && 94.0 && 66.8 \\
& & \multicolumn{1}{l}{}                                & Contrastive && 69.9 && 73.1 && 66.6 & 70.7 & \textbf{63.5} && 55.5 && 73.6 & 57.3 && 49.6 & 46.2 && \textbf{94.5} && 65.5 \\
& & \multicolumn{1}{l}{}                                & None && 70.7 && 65.7 && 66.0 & 69.4 & 53.8 && 53.3 && \textbf{75.9} & 53.6 && 53.8 & 50.0 && 93.4 && 64.2  \\ \\ \hline \\                     

\multicolumn{1}{l}{\STAB{\multirow{4}{*}{\rotatebox[origin=c]{90}{Spanish}}}}
& \multicolumn{1}{l}{}       
& \multicolumn{1}{l}{}                                & Both && 84.2 && \textbf{73.0} && 69.2 & \textbf{75.1} & 64.4 && 58.9 && \textbf{80.2} & \textbf{64.6} && 44.8 & 40.4 && 91.9 && \textbf{67.9} \\
& & \multicolumn{1}{l}{}                                & Pearson && \textbf{85.1} && 72.3 && \textbf{69.6} & 73.2 & 61.2 && 58.3 && 78.9 & 63.6 && \textbf{47.0} & 40.4 && 91.2 && 67.4 \\
& & \multicolumn{1}{l}{}                                & Contrastive && 74.7 && 72.7 && 66.8 &73.2 & \textbf{65.4} && 56.6 && 78.7 & 61.7 & &41.9 & 38.5 && \textbf{92.9} && 65.7 \\
& & \multicolumn{1}{l}{}                                & None && 75.8 && 66.5 && 68.2 & 67.2 & 61.8 && \textbf{61.3} && 79.0 & 61.5 && 42.9 & \textbf{46.2} && 91.1 && 65.6 \\ \\ \hline \\
                             
\multicolumn{1}{l}{\STAB{\multirow{4}{*}{\rotatebox[origin=c]{90}{Chinese}}}}
& \multicolumn{1}{l}{}    
& \multicolumn{1}{l}{}                                & Both && 78.0 && \textbf{69.3} && \textbf{69.0} & \textbf{72.2} & 65.4 && \textbf{55.7} && 74.1 & 63.8 && 48.0 & 49.0 && 91.6 && 66.9  \\
& & \multicolumn{1}{l}{}                                & Pearson && \textbf{82.0} && 68.5 && 68.3 & \textbf{72.2} & 63.4 && 52.7 && \textbf{75.2} & \textbf{64.7} && 49.2 & \textbf{53.8} && 90.8& & \textbf{67.3} \\
& & \multicolumn{1}{l}{}                                & Contrastive && 69.1 && \textbf{69.3} && 68.4 & 70.2 & \textbf{66.4} && 53.8 && 73.6 & 59.5 && 48.4 & 50.0 && \textbf{92.2} && 65.6 \\
& & \multicolumn{1}{l}{}                                & None && 69.1 && 66.4 && 66.8 & 69.3 & 63.0 && 55.1 && 75.9 & 62.2 && 46.3 & 41.3 && 88.3& & 64.0  \\ \\ \hline \\                               

\multicolumn{1}{l}{\STAB{\multirow{4}{*}{\rotatebox[origin=c]{90}{Portuguese}}}}
& \multicolumn{1}{l}{}         
& \multicolumn{1}{l}{}                                & Both && 78.8 && \textbf{74.3} && \textbf{70.2} & \textbf{74.2} & 63.1 && 54.4 && \textbf{79.0} & 64.1 && 50.7 & 56.7 && \textbf{93.7} && \textbf{69.0} \\
& & \multicolumn{1}{l}{}                                & Pearson && \textbf{80.4} && 71.0 && 68.0 & 72.1 & 55.9 && 51.8 && 78.5 & \textbf{64.5} && 47.2 & \textbf{57.7} && 92.7 && 67.2 \\
& & \multicolumn{1}{l}{}                                & Contrastive && 68.3 && 72.6 && 65.7 & 71.7 & 62.5 && 53.3 && 76.2 & 60.5 && 50.8 & 54.8 && 93.6 && 66.4 \\
& & \multicolumn{1}{l}{}                                & None  && 68.5 && 67.6 && 68.3 & 71.2 & 62.1 && 56.1 && 77.2 & 58.3 && 50.6 & 48.1 && 91.9 && 65.4 \\ \\ \hline \\

\multicolumn{1}{l}{\STAB{\multirow{4}{*}{\rotatebox[origin=c]{90}{Italian}}}}
& \multicolumn{1}{l}{}       
& \multicolumn{1}{l}{}                                & Both && 77.2 && 72.3 && \textbf{68.5} & \textbf{75.0} & 57.9 && 56.6 && 77.5 & 64.0 && 43.5 & 39.4 && 93.5 && 66.0 \\
 & & \multicolumn{1}{l}{}                                & Pearson && \textbf{77.8} && 71.2 && 68.2 & 73.1 & 52.7 && 56.4 && 77.8 & \textbf{64.8} && 45.5 & 45.2 && 93.1 && \textbf{66.0} \\
 & & \multicolumn{1}{l}{}                                & Contrastive && 61.4 && \textbf{72.6} && 64.5 & 71.7 & 60.1 && 56.8 && 77.5 & 61.2 && 46.9 & 44.2 && \textbf{94.2} && 64.6 \\
 & & \multicolumn{1}{l}{}                                & None && 67.3 && 67.5 && 65.9 & 68.2 & 52.1 && \textbf{58.7} && \textbf{78.4} & 62.7 && 46.5 & 50.0 && 91.3 && 64.4  \\ \\ \hline \\       

\multicolumn{1}{l}{\STAB{\multirow{4}{*}{\rotatebox[origin=c]{90}{Russian}}}}
& \multicolumn{1}{l}{}            
& \multicolumn{1}{l}{}                               & Both && \textbf{79.2} && \textbf{74.4} && \textbf{67.6} & \textbf{73.3} & \textbf{62.4} && \textbf{57.0} && \textbf{74.8} & 62.7 && 47.9 & \textbf{47.1} && 91.3 && \textbf{67.1} \\
& & \multicolumn{1}{l}{}                                & Pearson && 78.6 && 72.3 && 65.4 & 71.3 & 57.5 && 55.0 && 74.4 & \textbf{63.4} && 45.8 & 40.4 && 91.2 && 65.0 \\
 & & \multicolumn{1}{l}{}                               & Contrastive && 65.5 && 74.1 && 65.6 & 70.3 & 61.8 && 55.3 && 74.1 & 59.6 && \textbf{50.4} & 42.3 && \textbf{93.1} && 64.8 \\
& & \multicolumn{1}{l}{}                                & None && 70.5 && 65.8 && 66.0 & 68.0 & 56.2 && 53.6 && 72.2 & 58.3 && 49.9 & 49.0 && 88.3 && 63.4  \\ \\ \hline \\

\multicolumn{1}{l}{\STAB{\multirow{4}{*}{\rotatebox[origin=c]{90}{Korean}}}}
& \multicolumn{1}{l}{}       
& \multicolumn{1}{l}{}                               & Both && 61.0 && \textbf{69.4}& & 64.9 & \textbf{70.9} & 59.9 && 54.4 && 69.0 & 57.5 && 36.4 & 32.7 && 90.9 && \textbf{60.6} \\
& & \multicolumn{1}{l}{}                                & Pearson && 63.2 && 67.1 && \textbf{65.3} & \textbf{70.9} & 54.4 && 52.9 && 67.9 & 58.4 && 37.4 & 26.0 && 90.0 && 59.4 \\
& & \multicolumn{1}{l}{}                                & Contrastive && 48.5 && \textbf{69.4} && 63.8 & 68.2 & \textbf{61.9} && 52.1 && \textbf{69.1} & 55.5 && 35.6 & 35.6 && \textbf{91.5} && 59.2 \\
& & \multicolumn{1}{l}{}                                & None && 56.4 && 61.1 && 63.9 & 66.8 & 54.4 && \textbf{57.4} && \textbf{69.1} & 49.2 && \textbf{39.3} & 38.5 && 86.3 && 58.4  \\ \\ \hline \\                               

\multicolumn{1}{l}{\STAB{\multirow{4}{*}{\rotatebox[origin=c]{90}{Dutch}}}}
& \multicolumn{1}{l}{}        
& \multicolumn{1}{l}{}                                & Both& & 78.2 && \textbf{72.3} && \textbf{68.5} & \textbf{74.6} & 61.2 && 53.3 && 75.5 & 60.4 && 48.3 & 42.3 && \textbf{93.6} && \textbf{66.2} \\
& & \multicolumn{1}{l}{}                                & Pearson && \textbf{79.0} && 70.3 && 66.6 & 72.7 & 56.0 && 53.3 && \textbf{75.9} & \textbf{61.7} && 49.3 & 46.2 && 93.1 && 65.8 \\
& & \multicolumn{1}{l}{}                                & Contrastive& & 63.8 && 72.2 && 67.6 & 72.8 & \textbf{61.9} && 52.9 && 75.0 & 58.3 && 44.6 & 38.5 && \textbf{93.6} && 63.7 \\
& & \multicolumn{1}{l}{}                                & None && 64.4 && 66.5 && 62.8 & 67.2 & 49.2 && 55.1 && \textbf{75.9} & 59.2 && 46.6 & 47.1 && 92.9 && 62.5  \\ \\ \hline \\

\multicolumn{1}{l}{\STAB{\multirow{4}{*}{\rotatebox[origin=c]{90}{Average }}}}
& \multicolumn{1}{l}{}     
& \multicolumn{1}{l}{}                                  & Both& & 78.8	&& 73.0 &&	68.5 &	74.2&	62.4&&	56.1&&	75.5&	62.1&&	48.0&	43.3&&	93.6&&	66.8 \\
& & \multicolumn{1}{l}{}                                & Pearson && 80.4&&	71.1&&	68.0&	72.7&	56.9&&	55.0&&	75.9&	62.5&&	47.0&	45.2&&	92.7&&	66.1 \\
& & \multicolumn{1}{l}{}                                & Contrastive& & 68.3&&	72.6&&	66.6	&71.7	&62.5&&	54.6&&	75.0&	59.2&&	48.4&	44.2&&	93.6&&	64.9 \\
& & \multicolumn{1}{l}{}                                & None && 69.3&&	66.5&&	66.8&	69.3&	56.6&&	55.1&&	75.9&	58.3&&	49.9&	46.2&&	91.9&&	64.0  \\ \\ \hline \\
                                
\end{tabular}
}
\caption{Performance of the \href{https://huggingface.co/laion/CLIP-ViT-B-32-xlm-roberta-base-laion5B-s13B-b90k}{laion/CLIP-ViT-B-32-xlm-roberta-base-laion5B-s13B-b90k} model on the multilingual VALSE benchmark according to different metrics. \textbf{sns.} Counting small numbers. \textbf{adv.} Counting adversarial. \textbf{repl.} Action replacement. \textbf{Sp.rel.} Spatial relations.}
\label{tab:valse-extra}
\end{table*}

\begin{table*}[t!]
\centering
\resizebox{2\columnwidth}{!}{%
\begin{tabular}{l c c c   c   c c c   c   c c c   c   c c c   c   c c c    c   c c c }
 & \multicolumn{11}{c}{\scriptsize Multilingual CLIP ViT-B/32} & ~ & \multicolumn{11}{c}{\scriptsize Multilingual CLIP ViT-H/14} \\ \cline{2-12} \cline{14-24}
 & \multicolumn{3}{c}{\scriptsize Contrastive} & ~ & \multicolumn{3}{c}{\scriptsize Pearson} & ~ & \multicolumn{3}{c}{\scriptsize Combined} & ~ & \multicolumn{3}{c}{\scriptsize Contrastive} & ~ & \multicolumn{3}{c}{\scriptsize Pearson} & ~ & \multicolumn{3}{c}{\scriptsize Combined} \\ 
 \cline{2-4} \cline{6-8} \cline{10-12} \cline{14-16} \cline{18-20} \cline{22-24}  
 & ~~$\tau_b$~~ & ~~$\tau_c$~~ & ~~$\rho$~~ &    & ~~$\tau_b$~~ & ~~$\tau_c$~~ & ~~$\rho$~~ &    & ~~$\tau_b$~~ & ~~$\tau_c$~~ & ~~$\rho$~~ &    & ~~$\tau_b$~~ & ~~$\tau_c$~~ & ~~$\rho$~~ &   & ~~$\tau_b$~~ & ~~$\tau_c$~~ & ~~$\rho$~~ &   & ~~$\tau_b$~~ & ~~$\tau_c$~~ & ~~$\rho$~~ \\  \hline 
ENG      & 65.8 & 72.2 & 81.5 &  & 66.7 & 73.2 & 82.3 &  & 67.2 & 73.5 & 82.6 &  & 68.6 & 75.0 & 83.8 &  & 64.5 & 70.6 & 80.6 &  & \textbf{68.7} & \textbf{75.4} & \textbf{84.1}      \\ \hline
GER      & 65.1 & 71.4 & 80.8 &  & 65.6 & 72.1 & 81.4 &  & 66.5 & 72.6 & 81.9 &  & \textbf{68.0} & 74.3 & 83.2 &  & 62.9 & 69.1 & 79.3 &  & \textbf{68.0} & \textbf{74.8} & \textbf{83.6}      \\
FRE      & 65.4 & 71.7 & 81.1 &  & 66.0 & 72.4 & 81.7 &  & 66.7 & 72.9 & 82.1 &  & \textbf{68.0} & 74.4 & 83.3 &  & 62.8 & 69.0 & 79.3 &  & \textbf{68.0} & \textbf{74.7} & \textbf{83.6}     \\
SPA      & 65.1 & 71.3 & 80.8 &  & 65.9 & 72.6 & 81.7 &  & 66.4 & 72.7 & 81.9 &  & \textbf{67.8} & 74.3 & 83.2 &  & 62.4 & 68.6 & 78.9 &  & \textbf{67.8} & \textbf{74.5} & \textbf{83.5}      \\
CHI      & 64.4 & 70.6 & 80.2 &  & 64.7 & 71.3 & 80.7 &  & 65.7 & 71.9 & 81.3 &  & \textbf{67.4} & 73.8 & 82.8 &  & 62.1 & 68.3 & 78.6 &  & \textbf{67.4} & \textbf{74.0} & \textbf{83.1}     \\ 
POR      & 65.2 & 71.5 & 80.9 &  & 65.7 & 72.1 & 81.4 &  & 66.5 & 72.7 & 81.9 &  & \textbf{67.9} & 74.3 & 83.2 &  & 62.7 & 68.9 & 79.2 &  & \textbf{67.9 }& \textbf{74.6} & \textbf{83.5}      \\
ITA      & 65.0 & 71.3 & 80.7 &  & 65.6 & 72.2 & 81.4 &  & 66.3 & 72.6 & 81.8 &  & \textbf{67.9} & 74.3 & 83.2 &  & 62.5 & 68.7 & 79.0 &  & \textbf{67.9} & \textbf{74.6} & \textbf{83.5}      \\
RUS      & 64.6 & 70.8 & 80.3 &  & 65.1 & 71.6 & 80.9 &  & 65.9 & 71.9 & 81.4 &  & 67.4 & 73.8 & 82.8 &  & 62.7 & 68.9 & 79.2 &  & \textbf{67.6} & \textbf{74.2} & \textbf{83.2}     \\
KOR      & 63.8 & 70.0 & 79.6 &  & 64.3 & 70.8 & 80.2 &  & 65.1 & 71.2 & 80.7 &  & 67.3 & 73.7 & 82.7 &  & 62.1 & 68.3 & 78.6 &  & \textbf{67.4} & \textbf{74.1} & \textbf{83.1}     \\
DUT      & 65.2 & 71.4 & 80.8 &  & 65.8 & 72.4 & 81.6 &  & 66.6 & 72.8 & 82.0 &  & 68.1 & 74.5 & 83.4 &  & 62.9 & 69.1 & 79.3 &  & \textbf{68.2} & \textbf{74.9} & \textbf{83.8}       \\ \hline 
AVG~~    & 65.0 & 71.2 & 80.7 &  & 65.5 & 72.1 & 81.3 &  & 66.2 & 72.5 & 81.8 &  & 67.8 & 74.2 & 83.2 &  & 62.8 & 68.9 & 79.2 &  & \textbf{67.9} & \textbf{74.6} & \textbf{83.5}      \\

\hline 
\end{tabular}
}
\caption{Correlation between multilingual CLIPScore values and human rankings, considering machine-translated versions of the VICR dataset into 9 languages besides the original English. The last row presents macro-averaged correlation results across all the languages (including English). \textbf{Bold} values signify the best score per language.}
\label{tab:multilingual-results-2}
\end{table*}

\end{document}